# Panoptic Segmentation: A Review

Omar Elharrouss[a,∗], Somaya Al-Maadeed[a], Nandhini Subramanian[a], Najmath Ottakath[a], Noor almaadeed[a] and Yassine Himeur[b]

[a]*Department of Computer Science and Engineering, Qatar University, Doha, Qatar*
[b]*Department of Electrical Engineering, Qatar University, Doha, Qatar*

## ABSTRACT

Image segmentation for video analysis plays an essential role in different research fields such as smart city, healthcare, computer vision and geoscience, and remote sensing applications. In this regard, a significant effort has been devoted recently to developing novel segmentation strategies; one of the latest outstanding achievements is panoptic segmentation. The latter has resulted from the fusion of semantic and instance segmentation. Explicitly, panoptic segmentation is currently under study to help gain a more nuanced knowledge of the image scenes for video surveillance, crowd counting, self-autonomous driving, medical image analysis, and a deeper understanding of the scenes in general. To that end, we present in this paper the first comprehensive review of existing panoptic segmentation methods to the best of the authors' knowledge. Accordingly, a well-defined taxonomy of existing panoptic techniques is performed based on the nature of the adopted algorithms, application scenarios, and primary objectives. Moreover, the use of panoptic segmentation for annotating new datasets by pseudo-labeling is discussed. Moving on, ablation studies are carried out to understand the panoptic methods from different perspectives. Moreover, evaluation metrics suitable for panoptic segmentation are discussed, and a comparison of the performance of existing solutions is provided to inform the state-of-the-art and identify their limitations and strengths. Lastly, the current challenges the subject technology faces and the future trends attracting considerable interest in the near future are elaborated, which can be a starting point for the upcoming research studies. The papers provided with code are available at: https://github.com/elharroussomar/Awesome-Panoptic-Segmentation

## 1. Introduction

Nowadays, cameras, radars, Light Detection and Ranging (LiDAR), and sensors that capture data are highly prevalent [1]. They are deployed in smart cities to enable the collection of data from multiple sources in real-time, and be alerted to incidents as soon as they occur [2],[3]. They are also installed in public and residential buildings for security purposes. Therefore, there is a significant increase in the use of devices with video capturing capabilities. This leads to opportunities for analysis and inference through computer vision technology [4, 5]. This field of study has shot up in need due to the massive amounts of data generated from these equipment and the Artificial Intelligence (AI) tools that have revolutionized computing, e.g. machine learning (ML) and deep learning (DL), especially convolutional neural networks (CNNs). Videos and images captured contain useful information that can be used for several smart city applications, such as public security using video surveillance [6, 7], motion tracking [8], pedestrian behavior analysis [9, 10], healthcare services and medical video analysis [11, 12], and autonomous driving [13, 14], etc. Current needs and research trends in this field encourage further development. ML and big data analytic tools play an essential role in this field. On the other hand, computer vision tasks, e.g. object detection, recognition, and classification, rely on feature extraction, labeling, and segmentation of captured videos or images predominantly in real-time [8, 15, 16]. There is a strong need for properly labeled data in the AI learning process where information can be extracted from the images for multiple inferences. The labeling of images strongly depends on the considered application. Bounding box labeling and Image segmentation are some of the ways that videos/images can be labeled, which makes the automatic labeling a subject of interest [17, 18].

Nowadays, computer vision techniques augmented the robustness and efficiency of different technologies in our lives by enabling the various systems to detect, classify, recognize, and segment the content of any captured scenes using cameras. The segmentation of homogeneous regions or shapes in a scene that has similar texture or structure, such as the countable entities or objects termed things, and the uncountable regions, such as sky and roads that are termed stuff, is achieved [19, 20]. Indeed, the content of the monitored scene is categorized into things and stuff. Therefore, many visual algorithms are dedicated to identifying "stuff" and "thing" and denote a clear division between stuff and things. To that end, semantic segmentation has been introduced for identification of this couple (things, stuff) [21]. In contrast, instance segmentation is the technique that processes just the things in the image/video where an object is detected and isolated with a bounding box, or a segmentation mask [22, 23, 24].

Generally, image segmentation is the process of labeling the image content [25, 26]. Instance segmentation and semantic segmentation are traditional approaches to current

∗Corresponding author
∗∗Principal corresponding author

elharrouss.omar@gmail.com (O. Elharrouss); s\_alali@qu.edu.qa (S. Al-Maadeed); nandhini.reborn@gmail.com (N. Subramanian); gonajmago@gmail.com (N. Ottakath); n.alali@qu.edu.qa (N. almaadeed); yassine.himeur@qu.edu.qa (Y. Himeur)
ORCID(s):

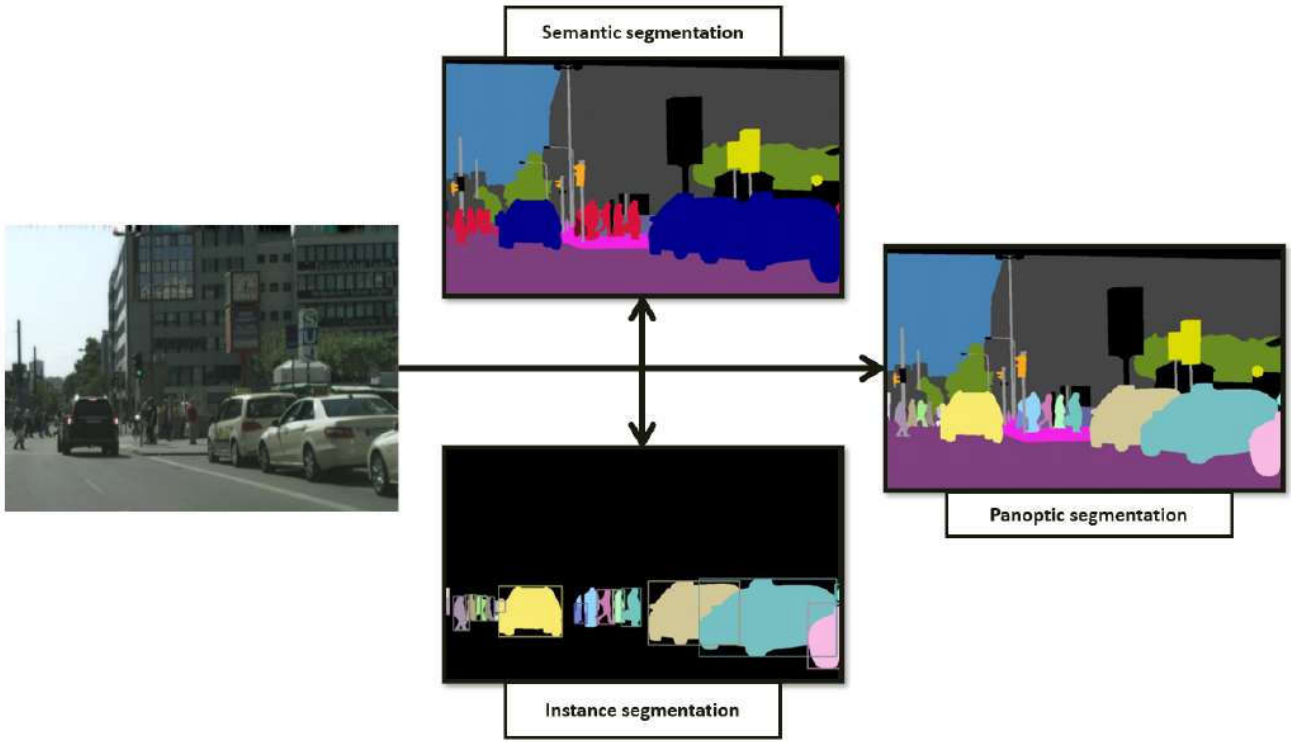

**Figure 1:** Sample segmentation results from [23] showing the difference among semantic segmentation, instance segmentation and panoptic segmentation.

trends in segmentation that mark each specific content in an image [27]. Semantic segmentation is the operation of labeling the things and the stuff by denoting the things of the same class with the same colors. While instance segmentation process the things in the image by labeling them with different colors and ignoring the background that can include the stuff [28, 29, 30]. In this line, objects can be further classified into "things" and "stuff," as in Kirillov et al., where stuff would be objects, such as the sky, water, roads, etc. On the other side, things would be the persons, cars, animals, etc. Put differently, in terms of segmentation; there is a high focus on identifying things and stuff separately, also separate (using colors) the things of the same class [31]. This operation is named panoptic segmentation, the developed version of segmentation that can include instance and semantic segmentation. Panoptic segmentation provides a clearer view of the image content, generates more information for its analysis, and enables computationally efficient operations using AI models by separating both things and stuff even of the same class or type using different colors. In this regard, Kirillov et al. have proposed the first framework of panoptic segmentation [23]. Panoptic segmentation can be done using the unified CNN-based method or by merging the instance and semantic segmentation results [32]. Fig. 1 portrays sample segmentation results from [23], that shows the difference between semantic segmentation, instance segmentation and panoptic segmentation.

Segmenting things and stuff in an image or video using panoptic segmentation is performed using different approaches, including unified models that are based on using panoptic segmentation masks and models that are based on combining stuff segmentation results and things segmentation results to obtain the final panoptic segmentation output [33]. For example, stuff classifiers may use dilated-CNNs, and on the flip side, object proposals are deployed, where a region-based identification task is often performed for object detection [34]. Even though each method has its own advantages, this isolative nature mostly can not reach the desired performance and/or may lack something significant and innovative that could be achieved in a combined way. This has led to a need for a bridge between stuff and things, where a detailed and coherent segmentation of a scene is generated within a combined framework of things and stuff. Accordingly, panoptic segmentation has been proposed as a well-worthy design, which can attain the aforementioned objectives [35]. This type of segmentation has great perspectives due to its efficiency and also as it can be used in a wide range of applications, including autonomous driving, augmented reality, medical image analysis, and remote sensing [36]. Even with the progress reached in panoptic segmentation, many challenges slow down the improvement in this field. These challenges are related to the scale of the objects in the scene, the complexity backgrounds in the monitored scenes, the cluttered scenes, the weather changes, the quality of used datasets, and the computational cost while using a large-scale dataset.

In this paper, we reviewed the existing panoptic segmentation techniques. This work sheds light on the most significant advances achieved on this innovative topic through conducting a comprehensive and critical overview of sate-of-the-art panoptic segmentation frameworks. Accordingly, it presents a set of contributions that can be summarized as follows:

- After presenting the background of the panoptic segmentation technology and its salient features, a thorough taxonomy of existing works is conducted concerning different aspects, such as the methodology deployed to design the panoptic segmentation model, types of image data that the subject technology and application scenarios could process.

- Public datasets deployed to validate the panoptic segmentation models are then discussed and compared with different parameters.

- Evaluation metrics are also described, and various comparisons of the most significant works identified in the state-of-the-art have been conducted to show their performance under different datasets and regarding various metrics.

- Current challenges that have been solved and those issues that remain unresolved are described before providing insights about the future directions attracting considerable research and development interest in the near and far future.

- Relevant findings and recommendations are finally drawn to improve the quality of panoptic segmentation strategies.

The remainder of the paper is organized as follows. Section 2 presents the background of the panoptic segmentation and its characteristics. Then, an extensive overview of existing panoptic segmentation frameworks is conducted in Section 3 based on a well-defined taxonomy. Next, public datasets are briefly described in Section 5 before presenting the evaluation metrics and various comparisons of the most significant panoptic segmentation frameworks identified in the state-of-the-art in Section 6. After that, current challenges and future trends are discussed in Section 7. Finally, the essential findings of this work are derived in Section 8.

## 2. Background

Image segmentation is an improvement on the object detection methods, in which image segmentation methods can be divided into semantic segmentation and instance segmentation. Semantic segmentation is used to classify each pixel in the image, whereas instance segmentation classifies each object in the scene. A detailed description of different types of image segmentation methods is given in this section.

### 2.1. Semantic Segmentation

Semantic segmentation can be defined as the pixel-level segmentation of the scenes in which a dense prediction is carried out. Put differently; semantic segmentation is the operation of labeling each pixel of an image with the corresponding class that represents the category of the pixel. Moreover, semantic segmentation classifies different regions in the images belonging to the same category of things or stuff. Even though semantic segmentation was proposed for the first time in 2007 when it became part of the computer vision, the significant breakthrough started once fully CNNs were first utilized by Long et al. [37] in 2014 for performing end-to-end segmentation of natural images.

For image segmentation, spatial analysis is the primary process that browses the image regions to decide the label of each pixel. CNN-based methods, such as U-Net, SegNet, fully connected networks (FCN), and DeconvNet, which are the basic architectures, succeed in segmenting these regions with acceptable accuracy in terms of the segmentation quality. However, for semantic segmentation, which is a complex segmentation, especially when the image is complex, the performance of these basic networks is not enough for labeling a large number of objects in the image. For example, the SegNet network heavily depends on the encoder-decoder architecture. In contrast, the other networks have similar architectures on the encoder side and only differ slightly in the decoder part of the architecture. To handle the problem of information lose, recent semantic segmentation methods that exploit deep convolutional feature extraction have been proposed using either multi-scale feature aggregation [38, 39, 40, 41] or end-to-end structured prediction perspectives [42, 43, 44, 45, 46].

### 2.2. Instance Segmentation

Instance segmentation is the incremental research work done based on the object detection task. The object (things) detection task not only detects the object but also gives a bounding box around the detected object to indicate the location [47]. Image segmentation is another step further to object detection, which segments the objects in the scene on a fine level and gives the label for all the objects in the segmented scene. The evolution order can be delivered as image classification, object detection, object localization, semantic segmentation, and instance segmentation. Segmentation efficiency refers to the computation time and cost, while accuracy refers to the ability to segment the objects correctly with robustness. Consequently, there is always a trade-off between accuracy and efficiency.

For any computer vision method, the selection of distinguishable features is of paramount importance, as the features are the crucial factor that decides the method's performance. Feature extractors, such as SIFT and SURF, were initially used before the introduction of the AI. Moving on, feature extraction slowly evolved from hand-picked manual methods to fully automated DL architectures. Some of the popular DL networks used for object detection are VGGNet [48], ResNet [49, 50], DenseNet [51, 52, 53], GoogLeNet

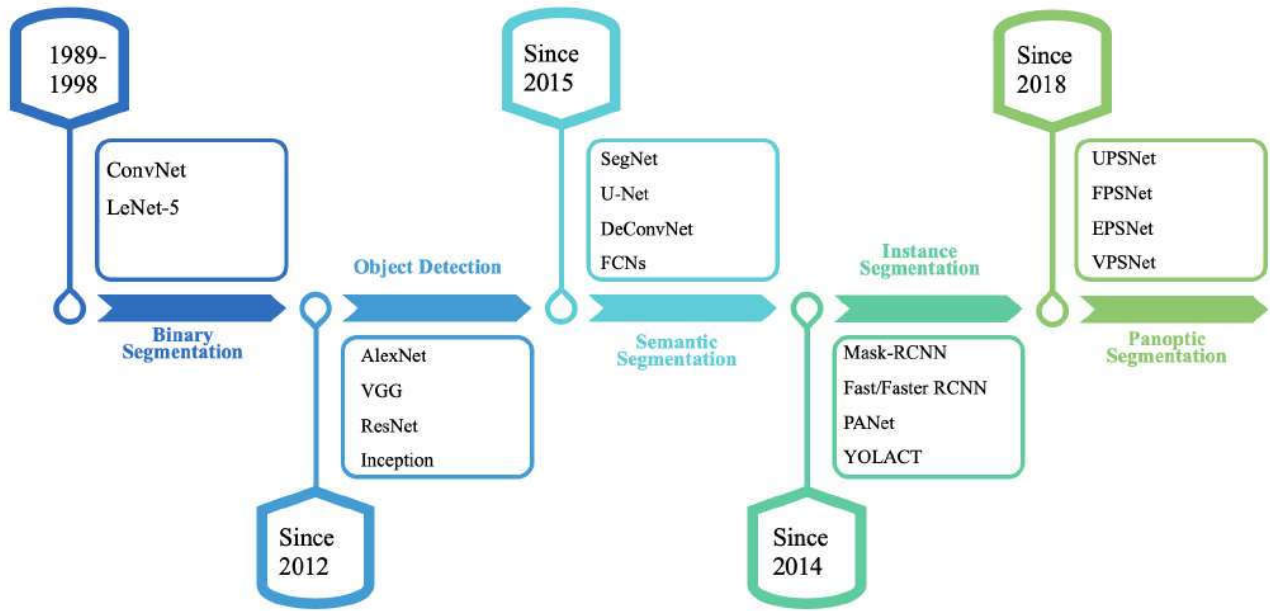

**Figure 2:** Timeline evolution of image segmentation.

[54], AlexNet [55, 56] , OverFeat [57], ZFNet [58], and Inception [59, 60]. In this context, the CNN architecture has been used as the backbone to extract features in some methods, which can be used for further processing. Besides, instance segmentation has to overcome several issues, including the geometric transformation, detection of smaller objects, occlusions, noises and image degradation. Accordingly, widely used architectures for instance segmentation include Mask-RCNN [61], RCNN [62, 63], path aggressive net (PANet) [64], and YOLACT [65, 66].

In general, instance segmentation is performed using either region-based two-stage methods [67, 68, 69, 39, 70, 71] or unified single-stage methods [72]. As discussed earlier, there is always a compromise between efficiency and accuracy. Two-stage methods have better accuracy, and single-stage methods have better efficiency. Unlike semantic segmentation, each object is differentiated from others even if they belong to the same class.

### 2.3. Panoptic segmentation

Panoptic segmentation is the fusion of the instance and semantic segmentation, which aims at differentiating between the things and stuff in a scene. Indeed, there are two categories in panoptic segmentation, i.e. stuff and things. Stuff refers to the uncountable regions, such as sky, pavements, and grounds. While things include all the countable objects, such as cars, people, and threes. Stuff and things are segmented in the panoptic method by giving each one of them a different color that distinguishes it from the others, unlike instance segmentation and semantic approaches, where we can find an overlapping between objects of the same type. Further, panoptic segmentation allows good visualization of different scene components and can be presented as a global technique that comprises detection, localization, and categorization of various scene parts. This leads to a bright and practical scene understanding.

The ability of panoptic segmentation techniques to describe the scene's content of an image and allow its deep understanding helps in significantly simplifying the analysis, improving the performance, and providing solutions to many computer vision tasks. We can find video surveillance, autonomous driving, medical image analysis, image scene parsing, geoscience, and remote sensing, among these tasks. Panoptic segmentation permits these applications by enabling the analysis of the specific targets without inspecting the entire regions of the image, which reduces the computational time, minimizes the miss-detection or recognition of some objects, and determines the edge saliency of different regions in an image or video. To examine the development of panoptic segmentation regarding the related tasks performed on things and stuff, a diagram that illustrates the timeline evolution of image segmentation starting from binary segmentation and object detection and ending by panoptic segmentation is described in Fig. 2. Typically, the popular networks used for performing each task have been highlighted as well.

## 3. Overview of panoptic segmentation techniques

Panoptic segmentation has been a breakthrough in computer vision; it enables the combinatorial view of "things" and "stuff". Thus, it represents a new direction in image segmentation. To inform the state-of-the-art, existing panoptic segmentation studies proposed in the literature are presented and deeply discussed in this section.

Some Panoptic segmentation techniques exploit instance and semantic segmentation separately before combining or aggregating the results to result in the panoptic segmentation. As a consequence, the shared backbone is used by taking the features generated by the backbone to be used in other parts of the network, as it is illustrated in Fig. 3(a). Other frameworks have used the same approaches but using an explicit connection between instance and semantic networks [73], as is portrayed Fig. 3(b).

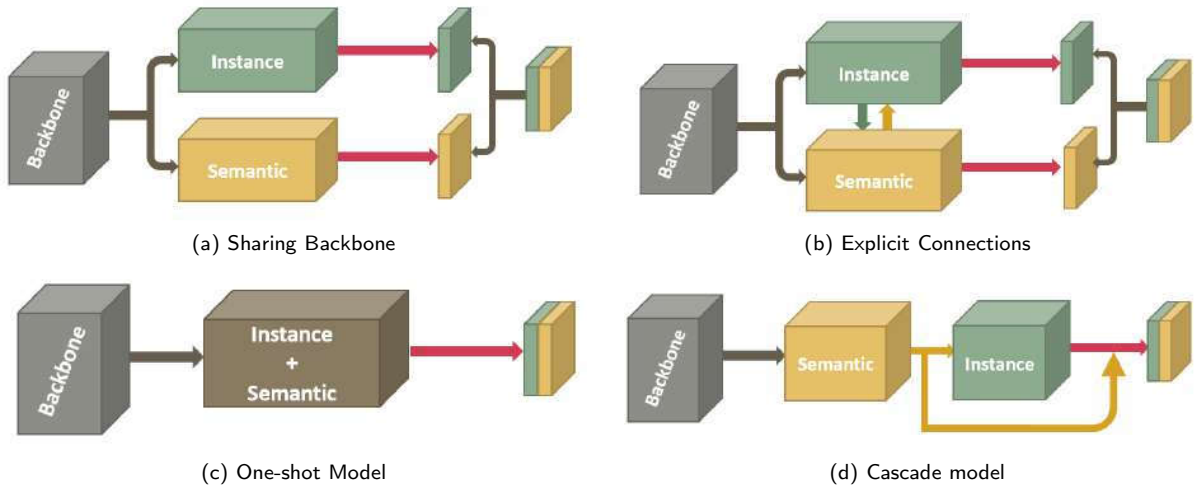

(a) Sharing Backbone

(b) Explicit Connections

(c) One-shot Model

(d) Cascade model

**Figure 3:** Networks methodologies for panoptic segmentation methods.

Most of the proposed panoptic segmentation frameworks used RGB images, while others performed their methods on medical images and LIDAR data. In this section, we discuss existing frameworks based on the type of data used.

### 3.1. RGB images data

RGB images represent the primary data source in which most of the panoptic segmentation algorithms have been applied. This is due to the wide use of RGB images in video cameras, image scanners, digital cameras, and computer and mobile phone displays. Also, the most proposed panoptic segmentation methods are performed on RGB images. For example, in [74], a panoptic segmentation model called Panoptic-Fusion is proposed, which is an online volumetric semantic mapping system combining both stuff and things. Aiming to predict class labels of background regions (stuff) and individually segment arbitrary foreground objects (things), it relies on first predicting pixel-wise panoptic labels for incoming RGB frames by fusing semantic and instance segmentation outputs. Similarly, in [75], Faraz et al. focus on improving the generalization abilities of networks to predict per-pixel depth from monocular RGB input images. A plethora of other panoptic methods have been designed to segment RGB images, such as [23, 31, 76, 77, 78].

To segment an image with the panoptic strategy, many frameworks have been proposed by first exploiting instance and semantic segmentation before concatenating the results of each part to obtain the final panoptic segmentation results. For that, Kirillov et al. [23], perform instance segmentation and semantic segmentation separately and then combine them both to obtain panoptic segmentation. Following, the non-maximum suppression (NMS) procedure is performed to develop the panoptic quality (PQ) metric. Therefore, the non-overlapping instance segments are produced using the NMS-like procedure, combined with semantic segmentation of the image. Moving on, slightly different modifications in the instance segmentation compared to [23] have been carried out in the following frameworks. In [43], the authors propose a CNN-based model using ResNet-50 backbone for features extraction followed by two branches, for instance and semantic segmentation. The two branches are merged using a heuristic method to come up with the final panoptic segmentation output. Moving on, to perform the panoptic segmentation and efficiently fuse the instances, an instance occlusion to Mask R-CNN is proposed in [79]. Typically, COCO and Cityscapes datasets were used to validate this scheme, which has an instance occlusion and other head for the Mask R-CNN that predicts occlusions between two masks. A good efficiency has been noted for multiple iterations, even for thousands, with the minimum overhead and addition of occlusion head on top of panoptic feature pyramid network (FPN). In the same context, in [80], another method, namely the pixel consensus voting (PCV), is used as a framework for panoptic segmentation. Indeed, PCV aims at elevating pixels to a first-class task, in which every pixel should provide evidence for the presence and location of objects it may belong to. As an alternative to region-based methods, panoptic segmentation based on pixel classification losses is simpler than other highly engineered state-of-the-art panoptic segmentation models.

In [81], the efficient spatial pyramid of dilated convolutions (ESPnet) is proposed, which has several stages to perform panoptic segmentation. Accordingly, a shared backbone consisting of FPN and ResNet is used along with a protohead generating prototypes. Following, both of them have been shared to instance segmentation and semantic segmentation heads. In this line, to enhance the input features, a cross-layer attenuation fusion module (CLA) is introduced, which aggregates the multi-layer feature maps in the FPN layer. This claims to outperform other approaches on the COCO dataset and has faster inference than most. In the same context, the authors in [82], propose a panoptic segmentation method, called EfficientPS, using the EfficientNet backbone as a multi-scale feature extraction scheme shared with semantic and instance modules. The results of each module are merged to obtain the final panoptic results.

Going further, in [83], the authors introduce a selection-oriented scheme using a generator system for guessing an ensemble of solutions and an evaluator system for ranking and selecting the best solution. In this case, the generator/evaluator method consists of two independent CNNs. The first one is used for guessing the segments corresponding to objects and regions in the image, and the second checks and selects the best components. Following, refinement and classification have been combined into a single modular system for panoptic segmentation. This approach works on a trial and error basis and has been tested on COCO panoptic dataset. In [84], the limitations of object reconstruction and pose estimation (RPE) are addressed by using panoptic mapping and object pose estimation system (Panoptic–MOPE). Thus, a fast semantic segmentation and registration weight prediction CNN using RGB-D data, namely fast-RGBD-SSWP, is deployed. Next, in [85], an FPN backbone is used in combination with Mask R-CNN for aggregating the instance segmentation method with a semantic segmentation branch, in which a shared computation has been achieved through this framework.

On the other hand, in [39], Panoptic-DeepLab is proposed, which is a simple design requiring only three loss functions during training. The Panoptic-DeepLab is the first bottom-up and single-shot panoptic segmentation that has attained state-of-the-art performance on public benchmarks, and thus it delivers end-to-end inference speed. Specifically, dual atrous spatial pyramid pooling (ASPP) for semantic segmentation and dual decoder, for instance, segmentation, have been deployed. In addition, the instance segmentation branch used is class agnostic, which involves a simple instance center regression, whereas the semantic segmentation design is a typical model. In [86], the background stuff and object instances (things) are retrieved from panoptic segmentation through an end-to-end single-shot method, which produces non-overlapping instance segmentation at a video frame rate. Therefore, merging outputs from sub-networks with instance-aware panoptic logits is considered and does not rely on instance segmentation. A notable PQ has been achieved through this method. In [87], an end-to-end occlusion-aware network (OANet) is introduced to perform a panoptic segmentation, which uses a single network to predict instance and semantic segmentation. Occlusion problems that may occur due to the predicted instances have been addressed using a spatial ranking module. In a similar fashion, in [88], an end-to-end trainable deep net using the bipartite conditional random fields (BCRF) combines the predictions of a semantic segmentation model and an instance segmentation model to obtain a consistent panoptic segmentation. A cross potential between the semantic and instance segmentation with the hypothesis the semantic label at any pixel has to be compatible with the instance label at that pixel has been introduced.

Using a Lintention-based network, the authors in [31] propose a two-stage-based panoptic segmentation method. Similar to the methods based on two separated networks [85], the LintentionNet architecture consists of an instance segmentation branch and semantic segmentation branch, where the fusion operation is introduced to produce the final panoptic results. In the same context, by exploring inter-task regularization (ITR) and inter-style regularization (ISR), the authors in [32] segment the images with a panoptic segmentation. Another two-stage panoptic segmentation method is proposed in [89]. Accordingly, after segmenting the instance using Mask-RCNN and the semantic using a DeepLab-based network, this model combines these results as a panoptic.

The object scale is one of the challenges for semantic, instance, and panoptic segmentation methods. The same object can be represented by a few pixels, taking a large region in the images. Thus, the segmentation of the objects with different scales affects the performance of methods. For that reason, Porzi et al. [90] propose a scale-based architecture for panoptic segmentation. While in [69], a deep panoptic segmentation based on a bi-directional learning pipeline is introduced. The authors train and evaluate their method using two backbones, including RestNet-50 and DCN-101. Intrinsic interaction between semantic segmentation and instance segmentation is modeled using a bidirectional aggregation network called BANet to perform a panoptic segmentation. Two modules have been deployed, instancing the context abundant features from semantic segmentation and instance segmentation for localization and recognition. Following, a feature aggregation is conducted through the bidirectional paths. Similarly, other cooperative methods of instance and semantic to perform panoptic segmentation are proposed in [73] and [91]. Typically, the model in [73] starts by a feature extraction process using shuffleNet, then performs the instance and semantic segmentation with explicit connections between the two-part. Finally, the obtained results have been fused to produce the panoptic segmentation output. While in [91], the authors introduce a two-stage panoptic segmentation scheme, which utilizes a shared backbone and FPN to extract multi-scale features. This scheme is extended with dual-decoders for learning background and foreground-specific masks. Finally, the panoptic head has assembled location-sensitive prototype masks based on a learned weighting approach with reference to the object proposals.

Fig. 4 illustrates four panoptic segmentation networks used , While Table 1 and 2 summarizes the backbones, characteristics and datasets used in each panoptic segmentation framework.

As described previously, some panoptic segmentation models generate segmentation masks by holding the information from the backbone to the final density map without any explicit connections. In this context, a panoptic edge detection (PED) is used to address a new finer-grained task, in which semantic-level boundaries for stuff classes are predicted along with instance-level boundaries for instance classes [93]. This provides a more comprehensive and unified understanding of the scene. Moving on, a panoptic edge network (PEN) brings together the stuff and instances into a single network with multiple branches. While in [70], the problem of low-fill rate linear objects and the inability to identify pixels near the bounding box has been taken into consideration. Thus, a trainable and branched multitask architecture has been used for grouping pixels for panoptic segmentation.

**Table 1**
Description of panoptic segmentation methods and the datasets used for their validation.

| Method | Backbone | Description | M. RCNN | Cityscape | COCO | M.Vistas | P. VOC | ADE20K |
|--------|----------|-------------|---------|-----------|------|----------|--------|--------|
| Kirillov et al. [23] | Self-design | Separate instance and semantic segmentation, merged using NMS and Heuristic method | ✓ | ✓ | ✗ | ✓ | ✗ | ✓ |
| LintentionNet [31] | ResNet-50-FPN | Efficient attention module (Lintention) | ✓ | ✗ | ✓ | ✗ | ✗ | ✗ |
| CVRN [32] | DeepLab-V2 | Exploits inter-style consistency and inter-task regularization modules | ✓ | ✓ | ✗ | ✗ | ✗ | ✗ |
| Panoptic-deeplab [39] | Xception-71 | Fusing predicted semantic, instance segmentation and instance regression | ✗ | ✓ | ✓ | ✓ | ✗ | ✗ |
| JSISNet [43] | ResNet-50 | Instance and semantic branches followed by a Heuristics merging | ✓ | ✗ | ✓ | ✗ | ✗ | ✗ |
| EfficientPS [82] | EfficientNet | Multi-scale features extraction, then merging semantic and instance head modules | ✓ | ✓ | ✗ | ✓ | ✗ | ✗ |
| EPSNet [81] | ResNet101-FPN | Merging instance and semantic segmentation with a cross-layer attention fusion | ✗ | ✗ | ✓ | ✗ | ✗ | ✗ |
| Zhang et al. [70] | FCN8 | Apply a clustering method on the output of merged instance and semantic branches | ✗ | ✓ | ✓ | ✗ | ✗ | ✗ |
| OCFusion [79] | ResNeXt-101 | Fusing instance and semantic branches and occlusion matrix outputs | ✓ | ✓ | ✓ | ✗ | ✗ | ✗ |
| PCV [80] | ResNet-50 | Instance segmentation by predicting objects centroids (voting map) merged with semantic results | ✓ | ✓ | ✓ | ✗ | ✗ | ✗ |
| Eppel et al. [83] | ResNet-50 | Generator-evaluator network for panoptic segmentation and class agnostic parts segmentation | ✗ | ✓ | ✗ | ✗ | ✗ | ✗ |
| Fast-RGBD-SSWP [84] | Self-design | RGB and depth images as inputs for 3D segmentation | ✓ | ✗ | ✗ | ✗ | ✗ | ✗ |
| Kirillov et al. [85] | ResNet-101 | Pyramid networks for instance and semantic branches | ✗ | ✓ | ✓ | ✗ | ✗ | ✗ |
| OANet [87] | ResNet50 | Spatial raking module for merging instance and semantic results | ✓ | ✗ | ✓ | ✗ | ✗ | ✗ |
| BANet [69] | ResNet-50-FPN, DCN-101-FPN | Explicit interaction between instance and semantic modules | ✓ | ✗ | ✓ | ✗ | ✗ | ✗ |
| Weber et al. [86] | ResNet50,FPN | Merging object detection, instance and semantic modules | ✗ | ✗ | ✗ | ✗ | ✗ | ✗ |
| Porzi et al. [90] | HRNetV2-W48+ | Merging instance and semantic branches with a crop-aware bounding box regression loss (CABB loss) | ✓ | ✓ | ✗ | ✓ | ✗ | ✗ |
| Auto-Panoptic [73] | shuffleNet | Merging instance and semantic branches with inter-modular fusion | ✓ | ✗ | ✓ | ✗ | ✗ | ✓ |
| Petrovai et al [91] | VoVNet2-39 | Multi-scale feature extraction shared with instance and semantic branches | ✗ | ✓ | ✗ | ✗ | ✗ | ✗ |
| Li et al. [92] | ResNet-101 | Weakly- and semi-supervised panoptic Segmentation by merging semantic and detector results | ✗ | ✓ | ✗ | ✗ | ✓ | ✗ |

Moving forward, a panoptic segmentation approach providing faster inference compared to existing methods is proposed in [46]. Explicitly, a unified framework for panoptic segmentation is utilized, which uses a backbone network and two lightweight heads for one-shot prediction of semantic and instance segmentation. Likewise, in [94], a mask R-CNN framework is used with a shared backbone named ResNet and FPN, which has individual network heads for each task. The created FPN provides multiscale information for the semantic segmentation, while the panoptic head learns to fuse the semantic segmentation logits with instance segmentation logit variables. Similarly, in [43], a single network with ResNet50 as the feature extractor is designed to (i) combine instance and semantic segmentation and (ii) produce the panoptic segmentation. Mask-RCNN is used as the backbone for instance segmentation, and the output is the classification score, class labels, and instance mask. The NMS heuristic is used to merge the results, classify and cluster the pixels to each class, and finally produce the panoptic image as the final result. Moving forward, in [95], another method is proposed for panoptic segmentation combining a multi-loss adaptation network.

On the other hand, in [67], a fast panoptic segmentation network (FPSNet) is proposed, which is faster than other panoptic methods as the instance segmentation and the merging heuristic part have been replaced with a CNN model called panoptic head. A feature map used to perform dense segmentation is used as the backbone for semantic segmentation. Moreover, the authors in [68] employ a position-sensitive attention layer which adds less computational cost instead of the panoptic head [68]. It utilizes a stand-alone method based on the use of a deep axial lab as the backbone. Also, a variant of the Mask R-CNN, as well as the KITTI panoptic dataset that has panoptic ground truth annotations, are proposed in [82, 75]. When compared with different approaches, this method has been found to be computationally efficient. In particular, a new semantic head has been used that coherently brings together nuanced and contextual features. The developed panoptic fusion model congruously integrates the output logits, in which the adopted pixel-wise classification has achieved a better picture of "things" and "stuff" (i.e. the countable and uncountable objects) in a given scene. Moreover, a lightweight panoptic segmentation and parametrized submodule that is powered by an end-to-end learned dense instance affinity is developed in [97], which captures the probability of pairs of pixels belonging to the same instance. The main advantage of this scheme is due to the fact that no post-processing is required. Furthermore, this method adds to the flexibility of the network design, in which its performance is noted to be efficient even when bounding boxes are used as localization queues. Similarly, a unified triple neural network is proposed in [71], which achieves more fine-grained segmentation. A shared backbone is deployed for extracting features of both things and stuff before fusing them. Here, stuff and things have been segmented separately, where the thing features are shared with stuff for performing a precise segmentation.

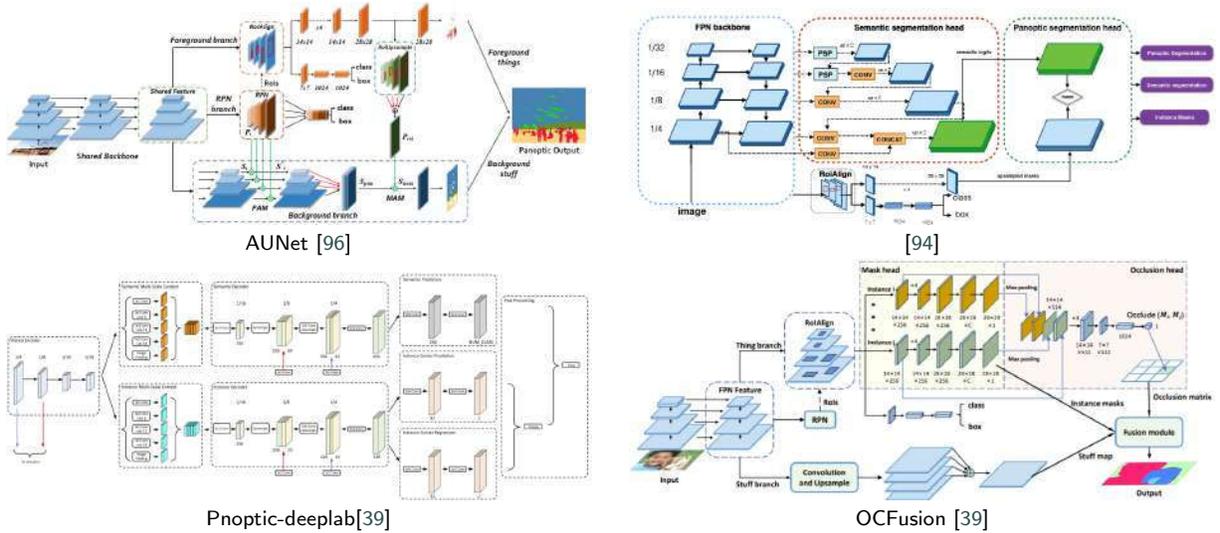

**Figure 4:** Example of panoptic segmentation networks

In [98], a new single-shot panoptic segmentation network is proposed, which performs real-time segmentation that leverages dense detection. Typically, a parameter-free mask construction method is used, which reduces the computation cost. Also, the information from object detection and semantic segmentation has been efficiently reutilized. Moreover, no feature map re-sampling is required in this simple data flow network design, which enables significant hardware acceleration. For the same objective, a single deep network is proposed in [40], which allows a more straightforward implementation on edge devices for street scene mapping in autonomous vehicles. Specifically, it had fewer computational costs with a factor of 2 when separate networks were used. The most likely and most reliable outputs of instance segmentation and semantic segmentation have then been fused to get a better PQ.

In [99], video panoptic segmentation (VPSnet), which is a new video extension of panoptic segmentation, is introduced, in which two types of video panoptic datasets have been used. The first refers to re-organizing the synthetic VIPER dataset into the video panoptic format for exploiting its large-scale pixel annotations. At the same time, the second has been built using the temporal extension on the Cityscapes val set via the production of new video panoptic annotations (Cityscapes-VPS). Moving on, the results were evaluated based on video panoptic metric (VPQ metric).

On the other side, a holistic understanding of an image in the panoptic segmentation task can be achieved by modeling the correlation between object and background. For this purpose, a bidirectional graph reasoning network for panoptic segmentation (BGRNet) is proposed in [100]. It adds a graph structure to the traditional panoptic segmentation network to identify the relation between things and background stuff. While in [101], with the spatial context of objects for both instance and semantic segmentation being a point of focus for interweaving features between tasks, a location-aware unified framework called spatial flow is put forth. Feature integration has facilitated different studies by making four parallel subnetworks.

Aiming at predicting consistent semantic segmentation, Porzi et al. use an FPN that produces contextual information from a CNN-based deep-lab module to generate multi-scale features [41]. Accordingly, this CNN architecture attains seamless scene segmentation results. Moving forward, whereas most of the existing works ignore modeling overlaps, overlap relations and resolving overlaps have been done with the scene overlap graph network (SOGnet) in [102]. This consists of joint segmentation, relational embedding module, overlap resolving module, and the panoptic head. While in [96], the foreground things and background stuff have been dealt together in attention guided unified network (AUNet). Typically, a unified framework has been used to achieve this combination simultaneously. Moreover, RPN and foreground masks have been added to the background branch, proving a consistency accuracy gain.

Without unifying the instance and semantic segmentation to get the panoptic segmentation, Hwang et al. [103] exploit the blocks and pathways integration that allows unified feature maps to represent the final panoptic outcome. In the same context, a unified method (DR1Mask) has been proposed in [104] based on a shared feature map of both instance and semantic segmentation for performing panoptic segmentation. Panoptic-DeepLab+ [105] has been proposed using squeeze-and-excitation and switchable atrous convolution (SAC) networks from SwideNet backbone to segment the object in a panoptic way.

From Semantic segmentation, the authors in [106] segment the instances of objects to generate the final panoptic segmentation. This method starts by segmenting the semantic using a CNN model then extracting the instance from the obtained semantic results. The panoptic segmentation is created using the concatenation between the results of each phase. In the same context and using instance-aware pixel embedding networks, a panoptic segmentation method is proposed in [107]. Also, a unified network named EffPS-b1bs4-RVC, a lightweight version of the EfficientPS architecture is introduced in [24]. While in [108], DetectoRS is implemented, which is a panoptic segmentation method that consists of two levels: macro level and micro level. At the macro level, a recursive feature pyramid (RFP) is used to incorporate different feedback connections from FPNs into the bottom-up backbone layers. While at the micro-level, a SAC is exploited to convolve the features with different atrous rates and gather the results using switch functions. Finally, in [109], aiming at visualizing the hidden enemies in a scene, a panoptic segmentation method is proposed.

## 3.2. Medical images

As medical imaging is being one of the most valued applications of computer vision, different kinds of images are used for both diagnosis and therapeutic purposes, such as X-rays, computed tomography (CT) scan, magnetic resonance imaging (MRI), ultrasound, nuclear medicine imaging and positron-emission tomography (PET). In this regard, medical image segmentation plays an essential role in computer-aided diagnosis systems. By assigning class values to each pixel and separating objects within the same class, instance segmentation is required. Typically, a unique ID is assigned to every single object. On the other hand, biological behaviors are studied and analyzed from the images' morphology, spatial location, and distribution of objects. As the instance segmentation has its limitations, cell R-CNN is proposed, which has a panoptic architecture. Typically, the encoder of the instance segmentation model is used to learn the global semantic-level features accomplished by jointly training a semantic segmentation model [110].

In [111], the focus was on histopathology images used for nuclei segmentation, where a CyC–PDAM architecture is proposed for this purpose. A baseline architecture is first designed, which performs an unsupervised domain adaptation (UDA) segmentation based on appearance, image, and instance-level adaption. Then a nuclei inpainting mechanism is designed to remove the auxiliary objects in the synthesized images, which is found to avoid false-negative predictions. Moving on, a semantic branch is introduced to adapt the features in terms of foreground and background, using semantic and instance-level adaptations, where the model learns the domain-invariant features at the panoptic level. Following, to reduce the bias, a re-weighting task is introduced. This scheme has been tested on three public datasets; it is found to outperform the-state-of-art UDA methods by a large margin. This method can be used in other applications with promising performance close to those of fully supervised schemes.

Moreover, the reader can refer to many other panoptic segmentation frameworks that have been developed to segment medical images and achieve different goals, such as pathology image analysis [112], prostate cancer detection [113] and segmentation of teeth in panoramic X-ray images [114].

## 3.3. LiDAR data

LiDAR is a technology similar to RADAR that can create high-resolution digital elevation models with vertical accuracy of almost 10 cm. LiDAR data is preferred for its accuracy and robustness, in which object detection [115, 116, 117, 118], and odometry [119, 120, 121] on the LiDAR space have already been carried out, and the focus has shifted towards panoptic segmentation of the LiDAR space. Accordingly, SematicKITTI dataset, an extension of KITTI dataset contains annotated LiDAR scans with different environments, and car scenes [122] has extensively been used. For example, in [123], two baseline approaches that combine semantic segmentation and 3D object detectors are used for panoptic segmentation. Similarly, in [124], Point-Pillars object detector is used to get the bounding boxes and classes for each object and a combination of KPConv [125] and RangeNet++ [126] is deployed to conduct instance segmentation of each class. The two baseline networks are trained and tested separately, and the results are merged in the final step to producing the panoptic segmentation. A hidden test set is then used to conduct an online evaluation of the LiDAR-based panoptic segmentation.

**Table 2**
Description of panoptic segmentation methods and the datasets used for their validation.

| Method | Backbone | Description | M. RCNN | Cityscape | COCO | M.Vistas | P. VOC | ADE20K |
|---|---|---|:---:|:---:|:---:|:---:|:---:|:---:|
| | | | | | | *Dataset* | | |
| FPSNet [67] | ResNet-50-FPN | Multi-scale feature extraction with a dense pixel-wise attention module for classification | ✓ | ✓ | ✗ | ✗ | ✗ | ✗ |
| Axial-DeepLab [68] | DeepLab | Unified networks with an axial-attention module | ✗ | ✓ | ✓ | ✓ | ✗ | ✗ |
| Li et al. [97] | ResNet50 | Object detection for instance segmentation merged with semantic results | ✗ | ✗ | ✗ | ✗ | ✗ | ✗ |
| Hou et al. [98] | ResNet-50-FPN | Single-shot network with a dense detection and a global self-attention module | ✗ | ✓ | ✓ | ✗ | ✗ | ✗ |
| VSPNet [99] | ResNet-50-FPN | Spatio-temporal attention network for video panoptic segmentation | ✗ | ✓ | ✗ | ✗ | ✗ | ✗ |
| BGRNet [100] | ResNet50,FPN | Explicit interaction between things and stuff modules | ✗ | ✗ | ✓ | ✗ | ✗ | ✓ |
| Geus et al. [40] | ResNet50 | Explicit interaction between instance and semantic modules then heuristic merging | ✗ | ✓ | ✗ | ✓ | ✗ | ✗ |
| PENet [93] | ResNet50 | Panoptic edge detection using object detection and instance and semantic segmentation | ✗ | ✗ | ✗ | ✗ | ✗ | ✓ |
| SpatialFlow[101] | ResNet50 | Location-aware and unified panoptic segmentation by bridging all the tasks using spatial information flows | ✗ | ✗ | ✓ | ✗ | ✗ | ✗ |
| SOGNet[102] | ResNet101,FPN | Scene graph representation for panoptic segmentation | ✗ | ✓ | ✓ | ✗ | ✗ | ✗ |
| AUNet [96] | ResNet50,FPN | Merging foreground and background branches with an attention-guided network | ✓ | ✓ | ✓ | ✗ | ✗ | ✗ |
| UPSNet [46] | ResNet-50-FPN | Sharing backbone by semantic and instance modules then applying a fusion module | ✓ | ✓ | ✓ | ✗ | ✗ | ✗ |
| Petrovai et al. [94] | ResNet50,FPN | FPN for panoptic segmentation | ✓ | ✓ | ✗ | ✗ | ✗ | ✗ |
| Saeedan et al. [75] | ResNet50 | An encoder-decoder network with a panoptic boost loss and geometry-based loss | ✗ | ✓ | ✗ | ✗ | ✗ | ✗ |
| SPINet [103] | ResNet50 | Panoptic-feature generator by integrating the execution flows | ✗ | ✓ | ✗ | ✗ | ✗ | ✗ |
| DR1Mask [104] | ResNet101 | Dynamic rank-1 convolution (DR1Conv) to merge high-level context information with low-level detailed features | ✗ | ✗ | ✓ | ✗ | ✗ | ✗ |
| Chennupati et al. [106] | ResNet50,101 | Instance contours for panoptic | ✗ | ✓ | ✗ | ✗ | ✗ | ✗ |
| Gao et al. [107] | ResNet101-FPN | category- and instance-aware pixel embedding (CIAE) to encode instance and semantic information | ✗ | ✓ | ✓ | ✗ | ✗ | ✗ |
| DetectoRS [108] | ResNeXt-101 | Backbone with a SAC module | ✗ | ✗ | ✓ | ✗ | ✗ | ✗ |
| Son et al. [109] | ResNet-50-FPN | Efficient panoptic network and completion network to reconstruct occluded parts | ✗ | ✓ | ✗ | ✗ | ✗ | ✗ |
| Ada-Segment [95] | ResNet-50 | Combining multi-loss adaptation networks | ✗ | ✗ | ✓ | ✓ | ✗ | ✓ |
| Panoptic-DeepLab+ [105] | SWideRNet | incorporating the squeeze-and-excitation and SAC to ResNets | ✗ | ✓ | ✓ | ✗ | ✗ | ✗ |

Moving on, when the CNN architecture is used frequently, a distinct and contrasting approach is adapted by Hahn et al. [127] to cluster the object segments. Since the clustering does not need computation time and energy as that of CNNs, the model adopted in [127] can be deployed even with a CPU. Nevertheless, the evaluation has been done on the SematicKITTI dataset, and the PQ, SQ, RQ and mIoU have been used as evaluation metrics. Going further, in [128], a Q-learning-based clustering method for panoptic segmentation of LiDAR point clouds is implemented by Gasperini et al., namely Panoster. While in [123], a two-stage approach is implemented based on combining LiDAR-based semantic segmentation and another detector that helps to enrich the segmentation with instance information. Besides, in [129], a unified approach is used by Milioto et al., where an end-to-end model is proposed. Specifically, the data is represented in range points, and features are extracted with a shared backbone. An image pyramid is used at the end of the backbone before using two decoders to reconstruct the panoptic image and the offsets' error estimation.

In [130], the authors propose PanopticTrackNet, which is a unified architecture used to perform semantic, instance segmentation, and object tracking. Accordingly, a common backbone called EfficientNet-B5 with FPN is used as the feature extractor. Thus, three separate modules, named semantic head, instance head, and instance tracking head, are used at the end of the backbone. Results from each of the modules are merged using a fusion module to produce the final result. Similar to PanopticTrackNet [130], the efficient LiDAR panoptic segmentation (EfficientLPS) [131] architecture has a unified 2D CNN model for the 3D LiDAR cloud data, except that there are some variations in the adopted backbone. EfficientLPS consists of a shared backbone comprising novel proximity convolution module (PCM), an encoder, the proposed range-aware FPN (RFPN), and the range encoder network (REN). A semantic head and an instance head are deployed at the end of the backbone to perform semantic and instance segmentation simultaneously. Finally, a panoptic fusion module is used to merge the results for panoptic segmentation. Moreover, a pseudo-label annotation is carried out to annotate the dataset for future use.

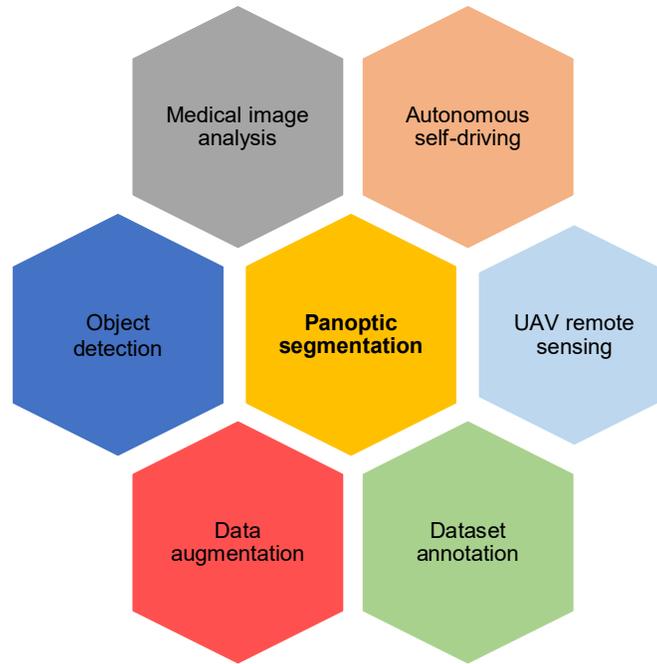

**Figure 5:** Applications of panoptic segmentation

In [132], a dynamic shifting network (DSNet) is deployed, which is a unified network with cylindrical convolutional backbone, semantic and instance head. Instead of the panoptic fusion, a consensus-driven fusion method is implemented. Moreover, a dynamic shifting method is employed to find the cluster centres of the LiDAR cloud points. To assign a semantic class and a temporally consistent instance ID to a sequence of 3D points, the authors in [133] develop a 4D panoptic LiDAR segmentation method. In this context, the semantic class for semantic segmentation has been determined, while probability distributions in the 4D spatio-temporal domain has been computed for the instance segmentation part.

## 4. Applications

The development of panoptic segmentation systems can be helpful for various tasks and applications. Thus, several case scenarios can be found where panoptic segmentation plays an essential role in increasing performance. Fig. 5 summarizes some of the main applications where panoptic segmentation has been involved.

### 4.1. Object detection

Panoptic segmentation has been mainly introduced for making the object detection process more manageable and accurate [23]. Object detection is a vital technology of computer vision and image processing. It refers to detecting instances of semantic objects of a particular class (e.g. humans, buildings, or cars) in digital images and videos. Panoptic segmentation has received significant attention for novel and robust object detection schemes [98, 134, 135].

### 4.2. Medical image analysis

The analysis and segmentation of medical images is a significant application based on segmenting objects of interest in medical images. Since the advent of panoptic segmentation, a great interest has been paid to using different panoptic models in the medical field [136]. For example, in [137], the issue of segmenting overlapped nuclei was taken into consideration, and a bending loss regularized network was proposed for nuclei segmentation. High penalties were reserved to contour with large curvatures, while minor curvatures were penalized with small penalties and used as a bending loss. This has helped minimize the bending loss and avoided the generated contours, which were surrounded by multiple nuclei. The MoNuSeg dataset was used to validate this framework using different metrics, including the aggregated Jaccard index (AJI), Dice, RQ, and PQ. This method claims to overperform other DL approaches using several public dataset. Moving forward, in [138], a panoptic feature fusion net is proposed, which is an instance segmentation process to analyze biological and biomedical images. Typically, the TCGA-Kumar dataset has been employed to validate the panoptic segmentation approach, which incorporates 30 histopathology images gathered from the cancer genome atlas (TCGA) [139].

### 4.3. Autonomous self-driving

Autonomous self-driving cars are a crucial application of panoptic segmentation. Scene understanding on a fine-grain level and a better perception of the scene is required to build an autonomous driving system effectively. Data collected from the hardware sensors, such as LiDAR, cameras, and radars, are crucial in enabling the possibilities of self-driving cars [140, 133, 141]. Moreover, the advance in DL and computer vision has led to an increased usage of sensor data for automation. In this context, panoptic segmentation can help accurately pars the images for both semantic content (where pixels represent cars vs. pedestrians vs. drivable space) and instance content (where pixels represent the same car vs. other car objects). Thus, planning and control modules can use panoptic segmentation output from the perception system for better informing autonomous driving decisions. For instance, the detailed object shape and silhouette information can help in improving object tracking, resulting in a more accurate input for both steering and acceleration. It can also be used in conjunction with dense (pixel-level) distance-to-object estimation methods to allow high-resolution 3D depth estimation of a scene. Typically, in [142], NVIDIA [1] develops an efficient scheme to perform pixel-level semantic and instance segmentation of camera images based on a single, multi-task learning DNN. This method has enabled the training of a panoptic segmentation-based DNN, which aims at understanding the scene as a whole versus piecewise. Accordingly, only one end-to-end DNN has been used for extracting all pertinent data while reaching per-frame inference times of approximately 5ms on an embedded in-car NVIDIA DRIVE AGX platform [2].

### 4.4. UAV remote sensing

Panoptic segmentation is an essential method for UAV remote sensing platforms, which can implement road condition monitoring and urban planning. Specifically, in recent years, the panoptic segmentation technology provides more comprehensive information than the current semantic segmentation technology [143]. For instance, in [144], the framework of the panoptic segmentation algorithm is designed for the UAV application scenario to solve some problems, i.e. the large target scene and small target of UAV, resulting in the lack of foreground targets of the segmentation results and the poor quality of the segmentation mask. Typically, a deformable convolution in the feature extraction network is introduced to improve network feature extraction ability. In addition, the MaskIoU module is developed and integrated into the instance segmentation branch for enhancing the overall quality of the foreground target mask. Moreover, a series of data are collected by UAV and organized into the UAV-OUC panoptic segmentation dataset to test and validate the panoptic segmentation models in aerial imagery [144].

### 4.5. Dataset annotation

Data annotation refers to categorizing and labeling data or images to validate the segmentation algorithms or other AI-based solutions. Panoptic segmentation can be also employed to perform dataset annotation [145, 146]. Typically, in [147], a panoptic segmentation is used to help conduct image annotation, which uses a collaborator (human) and automated assistant (based on panoptic segmentation) that both work together to annotate the dataset. The human annotator's action serves as a contextual signal for which the intelligent assistant reacts to and annotates other parts of the image. While in [92], a weakly supervised panoptic segmentation model is proposed for jointly doing instance segmentation and semantic segmentation and annotating datasets. This does not, however, detect overlapping instances. It has been tested on Pascal VOC 2012 that has up to 95% supervised performance. Moving on, in [76], an industrial application of panoptic segmentation for annotating datasets is studied. A 3D model is used to generate models of industrial buildings, which can improve inventories performed remotely, where a precise estimation of objects can be performed. For example, in a nuclear power plant site, the cost and time of maintenance can significantly be reduced since the equipment position can be first analyzed using panoptic segmentation of collected panoramic images before going on site. Therefore, this is presented as a huge breakpoint to advances in automation of large-scale industries using panoptic segmentation. In addition, a comprehensive virtual aeriaL image dataset, named VALID, is proposed in [143], that consists of 6690 high-resolution images that are annotated with panoptic segmentation and classified into 30 categories.

### 4.6. Data augmentation

Another promising application of panoptic segmentation is for data augmentation. By using panoptic segmentation, it becomes possible to design data augmentation schemes that operate exclusively in pixel space, and hence require no additional data or training, and are computationally inexpensive to implement [148, 149]. For instance, in [148], a panoptic data augmentation method, namely PandDA, is proposed. Specifically, the retraining of existing models of different PanDA augmented datasets (generated with a single frozen set of parameters), high-performance gains have been achieved in terms of instance segmentation and panoptic segmentation, in addition to detection across models backbones, dataset domains, and scales. Furthermore, because of the efficiency of unrealistic-looking training image datasets (synthesized by PanDA) there is emergence for rethinking the need for image realism to ensure a powerful and robust data augmentation.

### 4.7. Others

It is worth noting that panoptic segmentation can be used in other research fields, such as biology and agriculture, for analyzing and segmenting images. This is the case of [72], where panoptic segmentation is performed for behavioral research of pigs. This is although the evaluation does not directly affect the normal behavior of the animals, e.g., food and water consumption, littering, interaction, aggressive behavior, etc. Generally, object and Keypoint detectors are used to detect animals separately. However, the contours of the animals were not traced, which resulted in the loss of information. Panoptic segmentation has efficiently segmented individual pigs through a neural network (for semantic segmentation) using different network heads and post-processing methods to overcome this issue. The instance segmentation mask has been used to estimate the size or weight of animals. Even with dirty lenses and occlusion, the authors claim to have achieved 95% accuracy. Moreover, panoptic segmentation can be employed to visualize hidden enemies on battlefields, as described in [109].

---

[1] https://www.nvidia.com/en-us/

[2] https://www.nvidia.com/en-us/self-driving-cars/drive-platform/hardware/

## 5. Public data sources

Enabled by the growth of ML and DL algorithms, the datasets are the most important part of panoptic segmentation. With a dataset that contains thousands or billions of images with ground truth frames, there is a huge amount of data the models can learn from them. For example, ImageNet [150] helps to evaluate visual recognition algorithms, while VGGFace2 [151] helps to validate face recognition methods. For image segmentation, cityscape, Synthia, and Mapillary are the most famous. For panoptic segmentation, which is a new area of image segmentation with more details, only a few datasets have panoptic subjects. Fig. 6 portrays samples from the most famous datasets used to validate instance, semantic and panoptic segmentation techniques. In what follows, we briefly discuss the properties of these datasets.

**D1. Mapillary Vistas** [3]: It is an instance/semantic/panoptic segmentation dataset, which is represented as traffic-related dataset with a large-scale collection of segmented images [152]. This dataset is split into training, validation, and testing sets, while the size of each set is 18000, 2000, and 5000 images, respectively. The number of classes is 65 in total, where 28 refers to "stuff" classes and 37 is for "thing" classes. Moreover, it incorporates different image sizes ranged from $1024 \times 768$ to $4000 \times 6000$.

**D2. KITTI** [4]: The images pertaining to this dataset are captured from various places of the metropolis of Karlsruhe (Germany), including highways and rural regions. In each image, we can find around 30 pedestrians and 15 vehicles. Also, more than 12K images composed the KITTI dataset representing various objects. Thus, this dataset contains 5 different classes, where the number of "things" and "stuff" are not specified.

**D3. SemanticKITTI** [5]: It is another version of KITTI dataset that provides a 3D point-cloud version of the objects in KITTI dataset semanticKITTI. The data is captured with LiDAR from a field-of-view of 360 degrees. SemanticKITTI consists of 43 000 scans with 28 classes.

**D4. Middlebury Stereo** [6]: It is a dataset of depth images and RGB-D images captured from different field-of-view,

which are 7 with different object poses. It also contains many versions, including [153] and [154]. The images are gleaned with various resolutions including $690 \times 555$ and $1240 \times 1110$.

**D5. Cityscapes** [7]: It is the most used dataset, for instance semantic and panoptic segmentation tasks due to its large-scale size as well as the number of scenes represented in it, which have been captured from 50 cities [155]. The Cityscapes dataset is composed of more than 20K labeled images with coarse annotations. The number of object classes is about 30 classes.

**D6. COCO panoptic task** [8]: It is the most used dataset for image segmentation, and recognition [23]. For image segmentation, Microsoft COCO contains more than 2 million images while 328000 images are labeled for instance segmentation. Labeled images include stuff (various objects such as animals, people, etc.) and things (such as reads, sky, etc.). The novel version of COCO dataset assigned the instance and semantic labels for each pixel of any image with different colors, which is suitable for validating panoptic segmentation. For that, 123K images are labeled and divided into 172 classes, in which 91 of them refer to stuff classes and 80 for thing classes.

**D7. PSCAL VOC 2012** [9]: It has more than 11K images split into 20 classes. Also, this dataset contains stuff and things annotated, in which the stuff is labeled with the same colors [156].

**D8. ADE20K** [10]: It is another image segmentation dataset that includes about 25K images containing different types of objects where some of them are partially shown. This dataset is divided into the training set (20K), validation set (2K), and test set (3K). This repository also contains 50 stuff (sky and grass, etc.) and 100 things (cars, beds, person, etc.).

**D9. SYNTHIA** [11]: It is a synthetic dataset for scene segmentation scenarios in the context of driving scenarios [157]. The dataset is composed of 13 classes of things and stuff. While the number of things and stuff is not declared. SYNTHIA has more than 220K images with different resolutions also contains RGB and depth images.

**D10. TCGA-KUMAR**: It is a dataset of histopathology images for medical imagery purposes [139]. The dataset has 30 images of $1000 \times 1000$ pixels captured by medical sensors, e.g. TCGA at 40× magnification. The images represent many organs, such as the kidney, bladder, liver, prostate, and breast.

**D11. TNBC**: It is another histopathology dataset focusing on the triple-negative breast cancer (TNBC) dataset [158]. The TNBC dataset has 30 images with a resolution of $512 \times 512$ taken by the same sensors as TCGA-KUMAR dataset, and referred to 12 different patients. [158].

**D12. BBB** [12]: It is another medical imagery dataset of fluorescence microscopy images. It is composed of 200 images

---

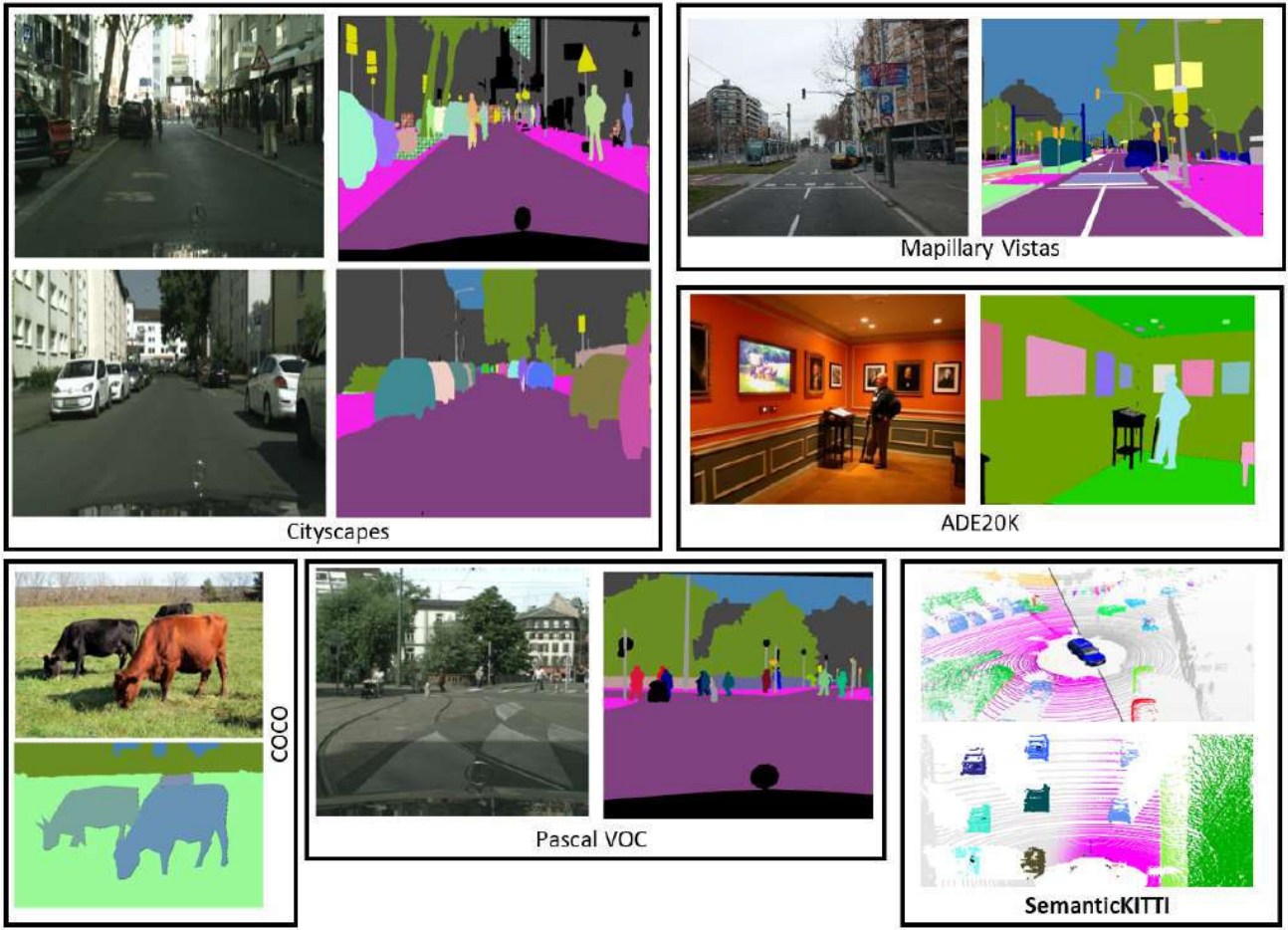

**Figure 6:** Samples from each dataset.

of a size of $520 \times 696$. The dataset has many versions, while the most used for panoptic segmentation is the BBBC039V1 dataset obtained from fluorescence microscopy. The dataset is divided into train set with 100 images, 50 images for validation, and 50 in the test set.

All in all, Table 3 summarizes the aforementioned datasets used for panoptic segmentation and their main properties, e.g. the number of images, resolution, attributes, geography/environment, format, and classes.

# 6. Result Analysis and discussion

In this section, we present the experimental results of the state-of-the-art works evaluated on standard panoptic segmentation benchmarks, including Cityscapes, COCO, Mapillary Vistas, Pascal VOC 2012, ADE20K, KITTI, and SemanticKITTI. The accuracy and the efficiency of each method are compared between all these methods. For each dataset, the methods are evaluated on val set and test set for each dataset, also this division is taken into consideration in this comparison.

## 6.1. Evaluation metrics (M)

Evaluation metrics are essential for informing the state-of-the-art and comparing the achieved results of any method against existing works. Panoptic segmentation, as discussed earlier, is a fusion of semantic and instance segmentation. Though the existing metrics for semantic and instance segmentation can suit panoptic segmentation to a certain extent, they can not be used exclusively. Thus, researchers have come up with different evaluation metrics to serve the purpose of panoptic segmentation. Generally, existing frameworks use panoptic quality (PQ), segmentation quality (SQ), and recognition quality (RQ) metrics for measuring the robustness of their results. In addition, other metrics are considered to compare the performance of such frameworks on the instance and semantic segmentation, such as the average over thing categories *th* and average over stuff categories *st*. Moreover, some methods also used the average precision (AP) and intersection-over-union (IoU) for comparison, which has been employed in this paper to compare existing methods. Overall, the evaluation metrics used for panoptic segmentation are given below.

**M1. Average Precision (AP):** To evaluate certain algorithms on the semantic, instance, and panoptic segmentation,

**Table 3**
Summary of the publicly available datasets for panoptic segmentation.

| Dataset | Images | Resolution | Attributes | Geography/Environment | Format | Classes |
|---|---|---|---|---|---|---|
| D1. Mapillary Vistas | 25000 | $4032 \times 3024$ | Real-World | Streets/ sun, rain, snow, fog .. | Images | 124(s), 100(T) |
| D2. KITTI Dataset | 12919 | $1392 \times 512$ | Real-World | Streets, Various objects | Videos | 5 |
| D3. SemanticKITTI | 43 000 | - | Real-World | Streets, Various objects | Point-wise | 28 |
| D4. Middlebury Stereo | 33 dataset | various | Real-World | Various objects | RGB-D Images | 33 |
| D5. Cityscapes | 5000 | $2040 \times 1016$ | Real-world | Streets/ climate(summer...) | Videos | 11 (S), 8 (T) |
| D6. COCO | 123,287 | Arbitrary | Real-world | Various objects | Images | 91(S),80(T) |
| D7. Pascal VOC 2012 | 11 530 | $2040 \times 1016$ | Real-world | Object & People in action | Videos | 20 |
| D8. ADE20K | 27 574 | $2040 \times 1016$ | Real-world | Various objects | Images | 100(T), 50(S) |
| D9. SYNTHIA | +200,000 | $1280 \times 760$ | Synthetic | misc, sky, building, road... | Videos | 13 |
| | +20,000 | $1280 \times 760$ | Synthetic | | Images | 13 |
| D10. TCGA-KUMAR | 30 | $1000 \times 1000$ | medical | U2OS cells | Images | - |
| D11. TNBC | 30 | $512 \times 512$ | medical | U2OS cells | Images | - |
| D12. BBBC039V1 | 200 | $520 \times 696$ | medical | U2OS cells | Images | - |

some metrics are used to report the quality of the segmented pixels simply. For example, we can find the PA that presents the percent of pixels in the image, correctly classified. Moreover, some works have also used mean average precision (mAP).

**M2. Intersection-over-union (IoU):** It also refers to the Jaccard index. It is essentially a method to quantify the percent overlap between the target mask and prediction output. This metric is closely related to the Dice coefficient, often used as a loss function during training. Quite simply, the IoU metric measures the number of pixels common between the target and prediction masks divided by the total number of pixels present across both masks.

$$IoU = \frac{target \cap prediction}{target \cup prediction} \qquad (1)$$

**M3. Panoptic quality (PQ):** It measures the quality in terms of the predicted panoptic segmentation compared to the ground truth involving segment matching and PQ computation from the matches. To make PQ insensitive to classes, PQ is calculated for individual classes, and then the average over all the classes is calculated. Three sets - true positive (TP), false positive (FP), and false-negative (FN) - and IoU between the ground truth and the prediction are required to calculate PQ. The expressions of SQ and RQ are given in Eqs. 2 and 3, respectively, while the formula for calculating PQ is given in Eq. 4.

$$SQ = \frac{\sum_{(p,q)\in TP} IoU(p,q)}{|TP|} \qquad (2)$$

where $p$ refers to the prediction and $g$ refers to the ground truth. The IoU between the prediction and the ground truth should be greater than 0.5 to give unique matching for the panoptic segmentation.

$$RQ = \frac{|TP|}{TP + \frac{1}{2}|FP| + \frac{1}{2}FN} \qquad (3)$$

The computation of PQ as defined in [23] and all panoptic segmentation methods is expressed as:

$$PQ = \frac{\sum_{(p,q)\in TP} IoU(p,q)}{TP + \frac{1}{2}|FP| + \frac{1}{2}FN} \qquad (4)$$

where $IoU(p,q)$ is the IoU between a predicted segment $p$ and the ground truth $g$. $TP$ refers to the matched pairs of segments, $FP$ denotes the unmatched predictions and $FN$ represents the unmatched ground truth segments. Additionally, $PQ^{Th}$, $SQ^{Th}$, $RQ^{Th}$ (average over thing categories), $PQ^{Sf}$, $SQ^{Sf}$, and $RQ^{Sf}$ (average over stuff categories) are reported to reflect the improvement on instance and semantic segmentation segmentation.

Finally, it is worthy to mention that the metrics described above have been introduced first by Kirillov et al. [23] and been adopted by other works as a common ground for comparison ever since, such as [82, 69, 147, 83]. Indeed, to evaluate the panoptic segmentation performance of such frameworks on medical histopathology and fluorescence microscopy images, AJI, object-level F1 score (F1), and PQ have been exploited.

**M4. Aggregated Jaccard index (AJI):** It is used for object-level segmentation measurement, also it is an extended version of Jaccard index and defined using the following expression:

$$AJI = \frac{\sum_{(p,q)} |G_i \cap P_M^i|}{\sum_{(p,q)} |G_i \cup P_M^i| + \sum |P_F|} \qquad (5)$$

where $G_i$ is the $ith$ nucleus in a ground truth with a total of $N$ nuclei. $U$ is the set of false positive predictions without the corresponding ground truth.

**M5. Object-level F1 score:** refers to the metric used for evaluating object detection performance, defined based on the number of true detection and false detection:

$$F1 = \frac{2TP}{FN + 2TP + FP} \qquad (6)$$

where TP, FN, and FP represent the number of true positive detection (corrected detected objects), false negative detection (ignored objects), and false positive detection (detected objects without corresponding ground truth), respectively.

## 6.2. Discussion

Image segmentation through panoptic segmentation has charted new directions to research in computer vision. Computers can now see as humans with a clear view of the scene. In this section, detailed analysis and discussion of the results obtained by existing frameworks are provided. To evaluate the performance of state-of-the-art models on different datasets, many metrics are used, including PQ, SQ, and RQ as well as on "thing" classes ($PQ^{th}$, $SQ^{th}$, and $RQ^{th}$) and "stuff" classes ($PQ^{st}$, $SQ^{st}$, and $RQ^{st}$). Most methods have provided the metrics on val set, while some approaches have been tested on both val set and test-dev set. For that reason, all the methods reported in this paper are collected and compared in terms of the data used in the test phase and metrics deployed to measure the performance. Furthermore, the results are grouped with reference to the dataset used for experimental validation.

Therefore, this article represents a solid reference to inform the state-of-the-art panoptic segmentation field, in which the actual research on panoptic segmentation strategies is carried out. This study produces a detailed comparison of the approaches used, datasets, and models adopted in each architecture and the evaluation metrics. Moving forward, the performance of existing methods has been clearly shown. With the methodology being the focus point, we can note that the architecture of panoptic segmentation methods varies with the approaches at different levels of the network. Although most frameworks use CNN as the base for their architectures, some have worked on attention models as well as mask transformers, either combined or standalone. The commonly used backbone for feature extraction is ResNet50 and ResNet101.

### 6.2.1. Evaluation on Cityscapes

Cityscapes is the most commonly preferred dataset for experimenting with the efficiency of panoptic segmentation solutions. A detailed report on the methods that use this dataset with the evaluation metrics is given in Table 4. In addition, the obtained results have been presented considering the dataset used for evaluation. Though it is common to use the val set for reporting the results, some works have reported their results on the test-dev set of the Cityscapes dataset. All the models are representative, and the results listed in Table 4 have been published in the reference documents. Moreover, all these works have been recently published in the last three years, such as WeaklySupervised (2018) [92], Panoptic-DeepLab (2019) [105], and EfficientPS (2020) [82].

The obtained results on test-dev set in Table 4 demonstrate that PanopticDeeplab+ [105], EfficientPS [82], and Axial-DeepLab [68] have reached the highest values in terms of PQ, where PanopticDeeplab+ outperforms EfficientPS and Axial-DeepLab by 0.7% and 2.2%, respectively. Also, it has been improved by more than 6% compared to other methods, e.g. SOGNet [102] and AUNet [96], which have used ResNet50-FPN as the backbone, while PanopticDeeplab+ and EfficientPS have utilized SWideRNet and

EfficientNet architectures as backbones, respectively. For the other metrics, such as SQ and RQ, EfficientPS [82] and Li et al. [97] have reached the best performance with 83.4% and 82.4% for SQ and 79.6% and 75.9% in terms of RQ. While Kirolov et al. [23] comes in third place for both SQ and RQ metrics. Besides, the other methods did not provide the obtained values for these two metrics. For the PQ of *thing* and *stuff* classes $PQ^{st}$ and $PQ^{th}$, the results achieved using EfficientPS [82] and Li et al. [97] are the highest in terms of $PQ^{st}$. While EfficientPS, SOGNet and Li et al. [97] have achieved the best $PQ^{th}$ results with a difference of 5.1% between EfficientPS and Li et al. [97] which comes in the third place with 54.8% for $PQ^{th}$.

On another side, it has been clearly shown that most of the frameworks have tested their methods on Cityscapes val set, unlike the test-dev set. Also, we can observe that different methods, such as EfficientPS [82], and Li et al. [97] have evaluated their results using the three metrics, i.e. PQ, SQ, and RQ. While some approaches have been evaluated using these metrics on things ($PQ^{th}$, $SQ^{th}$, and $RQ^{th}$) and stuff ($PQ^{st}$, $SQ^{st}$, and $RQ^{st}$), e.g. Geus et al. [40], UPSNet [46], and CASNet [160]. Moreover, from the results in Table 4, PanopticDeeplab+ reaches the highest PQ values, with an improvement of 2.6% than the second-best result obtained by Panoptic-deeplab, Axial-DeepLab, and CASNet. Regarding the RQ metric, EFFIcientPS achieves the best result using single-scale and multi-scale; it is better than Li et al.[97] by 5%. Using SQ metric, CASNet reaches 83.3%, which outperforms EfficientPS. In terms of $PQ^{th}$, $SQ^{th}$, and $RQ^{th}$, EfficientPS provides the best accuracy results. The difference between EfficientPS and other methods is that it utilizes a pre-trained model on Vistas dataset, where the schemes do not use any pre-training. In addition, EfficientPS uses EfficientNet as backbone while most discussed techniques have exploited RestNet50 except Panoptic-deeplab, which uses Xception-71 as a backbone.

### 6.2.2. Evaluation on COCO

The COCO dataset was originally introduced for object detection, which later extended for semantic, instance, and panoptic segmentation. COCO is the second preferred dataset after Cityscapes. Similar to the latter, both the test-dev and val set have been used for reporting the results. The average evaluation metrics values on the COCO dataset are not that high as Cityscapes. The reason could be that COCO is focused more on things since the dataset was initially developed for object detection.

Table 5 represents some of the obtained results using existing panoptic segmentation techniques. Similar to the performance presentation on Cityscapes, we present the results provided in different works for COCO, including those tested using test-dev set and val set. Using the performance results provided in the reference works, it has been obviously shown that for almost all the methods, the evaluation has been performed using all metrics, including PQ, SQ, and RQ, and the performance using the same metrics on the thing and stuff classes. From Table 4 and Table 5, we find

**Table 4**
Performance comparison of existing schemes on the val and test-dev sets under Cityscapes datasets, where red, blue, cyan colors indicates the three best results for each set.

| Dataset | test/val | Method | PQ | | | SQ | | | RQ | | |
|---|---|---|---|---|---|---|---|---|---|---|---|
| | | | PQ | PQ^st | PQ^th | SQ | SQ^st | SQ^th | RQ | RQ^st | RQ^th |
| Cityscapes | test | Kirillov et al. [23] | 61.2 | 66.4 | 54.0 | 80.9 | - | - | 74.4 | - | - |
| | | Axial-DeepLab [68] | 65.6 | - | - | - | - | - | - | - | - |
| | | SOGNet [102] | 60.0 | 62.5 | 56.7 | - | - | - | - | - | - |
| | | Zhang et al. [70] | 60.2 | 67.0 | 51.3 | - | - | - | - | - | - |
| | | Kirillov et al. [85] | 58.1 | 62.5 | 52 | - | - | - | - | - | - |
| | | AUNet [96] | 59.0 | 62.1 | 54.8 | - | - | - | - | - | - |
| | | EfficientPS [82] | 67.1 | 71.6 | 60.9 | 83.4 | | | 79.6 | | |
| | | Li et al. [97] | 63.3 | 68.5 | 56.0 | 82.4 | 83.4 | 81.0 | 75.9 | 80.9 | 69.1 |
| | | Panoptic-deeplab [39] | 65.5 | - | - | | | | | | |
| | | PanopticDeeplab+ [105] | 67.8 | - | - | | | | | | |
| | val | FPSNet [67] | 55.1 | 60.1 | 48.3 | - | - | - | - | - | - |
| | | Axial-DeepLab [68] | 67.7 | | | | | | | | |
| | | EfficientPS Single-scale[82] | 63.9 | 66.2 | 60.7 | 81.5 | 81.8 | 81.2 | 77.1 | 79.2 | 74.1 |
| | | EfficientPS Multi-scale [82] | 65.1 | 67.7 | 61.5 | 82.2 | 82.8 | 81.4 | 79.0 | 81.7 | 75.4 |
| | | Li et al. [97] | 61.4 | 66.3 | 54.7 | 81.1 | 81.8 | 80.0 | 74.7 | 79.4 | 68.2 |
| | | Panoptic-deeplab [39] | 67.0 | - | - | | | | | | |
| | | Hou et al. [98] | 58.8 | 63.7 | 52.1 | - | - | - | - | - | - |
| | | OCFusion [79] | 60.2 | 64.7 | 54.0 | - | - | - | - | - | - |
| | | Geus et al. [40] | 45.9 | 50.8 | 39.2 | 74.8 | - | - | 58.4 | - | - |
| | | PCV [80] | 54.2 | 58.9 | 47.8 | - | - | - | - | - | - |
| | | VPSNet [99] | 62.2 | 65.3 | 58.0 | - | - | - | - | - | - |
| | | SpatialFlow [101] | 58.6 | 61.4 | 54.9 | - | - | - | - | - | - |
| | | WeaklySupervised [92] | 47.3 | 52.9 | 39.6 | - | - | - | - | - | - |
| | | Porzi et al. [41] | 60.2 | 63.6 | 55.6 | - | - | - | - | - | - |
| | | UPSNet [46] | 61.8 | 64.8 | 57.6 | 81.3 | - | - | 74.8 | - | - |
| | | PanoNet [159] | 55.1 | - | - | - | - | - | - | - | - |
| | | CASNet [160] | 66.1 | 75.2 | 53.6 | 83.3 | - | - | 78.4 | - | - |
| | | PanopticDeepLab+ [105] | 69.6 | - | - | | | | | | |
| | | SPINet [103] | 63.0 | 67.3 | 57.0 | - | - | - | - | - | - |
| | | Chennupati et al. [106] | 48.4 | 59.3 | 33.2 | - | - | - | - | - | - |
| | | Son et al. [109] | 58.0 | - | - | 79.4 | - | - | 71.4 | - | - |
| | | Petrovai et al [91] | 57.3 | 62.4 | 50.4 | - | - | - | - | - | - |

that the used backbones have a crucial impact on the results, especially the deeper backbones, such as Xception-71 and RestNet-101. Accordingly, ResNeXt-101 has been used in Panoptic-deeplab [105], SOGNet [102], Ada-Segment[95] and DetectoRS [108]. For example, the Xception-71 backbone used as encoder for panoptic segmentation models, has 42.1 M parameters while RestNet-101 includes 44.5 M and ResNet-50 used in Ada-Segment[95] encompasses 25.6 M parameters. The use of deeper backbones demonstrates the effectiveness of SOGNet on test-dev set, which has utilized ResNet-101 and ResNeXt-101 backbones. It has also been observed that the results of various works in terms of PQ, RQ and SQ have a slight variation, such as BANet, AUNet, Dao et al. [161], UPSNet [46], Panoptic-DeepLab+ [105], Ada-Segment[95], and DetectoRS [108]. Moving on, the obtained results evaluated on things and stuff using the methods described in these references are the best. For example, BANet [69] has reached 82.1% and 66.3% regarding $RQ^{th}$ and $SQ^{th}$, respectively. While in terms of $PQ^{st}$, $SQ^{st}$ and

$RQ^{st}$ Geo et al. [161] has achieved the highest performance results with a rate of 51.8%, 81.4%, and 63%, respectively.

On the COCO val set, the evaluation results are slightly different from the obtained results on test-dev set, although some frameworks have reached the highest performance, such as Li et al. [97], OCFusion [111], Panoptic-DeepLab+ [105], OANet [87], and DR1Mask [104]. For example, using OCFusion, the performance rate for using PQ and $PQ^{th}$ metrics has reached 46.6% and 53.5%, respectively. The difference between the method that reaches the highest results and those in the second and third places is around 1-2%, which demonstrates the effectiveness of these panoptic segmentation schemes.

### 6.2.3. Evaluation on Mapillary Vistas

Mapillary Vistas has been used only by a few methods, where the average PQ value has attained 38.65%. The highest value of 44.3% has been reported by Panoptic-DeepLab+

**Table 5**
Performance comparison of existing schemes on the val and test-dev sets under COCO datasets, where red, blue, cyan colors indicates the three best results for each set.

| Dataset | test/val | Method | PQ | | | SQ | | | RQ | | |
|---|---|---|---|---|---|---|---|---|---|---|---|
| | | | PQ | PQ$^{st}$ | PQ$^{th}$ | SQ | SQ$^{st}$ | SQ$^{th}$ | RQ | RQ$^{st}$ | RQ$^{th}$ |
| COCO | test | JSISNet [43] | 27.2 | 23.4 | 29.6 | 71.9 | 72.3 | 71.6 | 35.9 | 30.6 | 39.4 |
| | | Axial-DeepLab [68] | 44.2 | 36.8 | 49.2 | - | - | - | - | - | - |
| | | EPSNet [81] | 38.9 | 31.0 | 44.1 | - | - | - | - | - | - |
| | | BANet [69] | 47.3 | 35.9 | 54.9 | 80.8 | 78.9 | 82.1 | 57.5 | 44.3 | 66.3 |
| | | OCFusion [79] | 46.7 | 35.7 | 54.0 | - | - | - | - | - | - |
| | | PCV [80] | 37.7 | 33.1 | 40.7 | 77.8 | 76.3 | 78.7 | 47.3 | 42.0 | 50.7 |
| | | Eppel et al. [83] | 33.7 | 31.5 | 35.1 | 79.6 | 78.4 | 80.4 | 41.4 | 39.3 | 42.9 |
| | | SpatialFlow [101] | 42.8 | 33.1 | 49.1 | 78.9 | - | - | 52.1 | - | - |
| | | SOGNet [102] | 47.8 | - | - | 80.7 | - | - | 57.6 | - | - |
| | | Kirillov et al. [85] | 40.9 | 29.7 | 48.3 | - | - | - | - | - | - |
| | | AUNet [96] | 46.5 | 32.5 | 55.8 | 81.0 | 77.0 | 83.7 | 56.1 | 40.7 | 66.3 |
| | | OANet [87] | 41.3 | 27.7 | 50.4 | - | - | - | - | - | - |
| | | UPSNet [46] | 46.6 | 36.7 | 53.2 | 80.5 | 78.9 | 81.5 | 56.9 | 45.3 | 64.6 |
| | | Panoptic-DeepLab+ [105] | 46.5 | 38.2 | 52.0 | - | - | - | - | - | - |
| | | Ada-Segment[95] | 48.5 | 37.6 | 55.7 | 81.8 | - | - | 58.2 | - | - |
| | | Gao et al. [107] | 46.3 | 37.9 | 51.8 | 80.4 | 78.8 | 81.4 | 56.6 | 46.9 | 63.0 |
| | | DetectoRS [108] | 50.0 | 37.2 | 58.5 | - | - | - | - | - | - |
| | val | JSISNet [43] | 26.9 | 23.3 | 29.3 | 72.4 | 73 | 72.1 | 35.7 | 30.4 | 39.2 |
| | | Axial-DeepLab [68] | 43.9 | 36.8 | 48.6 | - | - | - | - | - | - |
| | | EPSNet [81] | 38.6 | 31.3 | 43.5 | - | - | - | - | - | - |
| | | Li et al. [97] | 43.4 | 35.5 | 48.6 | 79.6 | - | - | 53.0 | - | - |
| | | BANet [69] | 43.0 | 31.8 | 50.5 | 79.0 | 75.9 | 81.1 | 52.8 | 39.4 | 61.5 |
| | | Hou et al. [98] | 37.1 | 31.3 | 41.0 | - | - | - | - | - | - |
| | | OCFusion [79] | 46.3 | 35.4 | 53.5 | - | - | - | - | - | - |
| | | PCV [80] | 37.5 | 33.7 | 40.0 | 77.7 | 76.5 | 78.4 | 47.2 | 42.9 | 50.0 |
| | | BGRNet [100] | 43.2 | 33.4 | 49.8 | - | - | - | - | - | - |
| | | SpatialFlow [101] | 40.9 | 31.9 | 46.8 | - | - | - | - | - | - |
| | | Weber et al. [86] | 32.4 | 28.6 | 34.8 | - | - | - | - | - | - |
| | | SOGNet [102] | 43.7 | 33.2 | 50.6 | - | - | - | - | - | - |
| | | OANet [87] | 40.7 | 26.6 | 50 | 78.2 | 72.5 | 82.0 | 49.6 | 34.5 | 59.7 |
| | | UPSNet [46] | 43.2 | 34.1 | 49.1 | 79.2 | - | - | 52.9 | - | - |
| | | Panoptic-DeepLab+ [105] | 45.8 | 38.0 | 51.0 | - | - | - | - | - | - |
| | | LintentionNet [31] | 41.4 | 30.8 | 48.3 | 78.5 | 72.9 | 82.2 | 50.5 | 39.1 | 58.0 |
| | | Ada-Segment[95] | 43.7 | 32.5 | 51.2 | - | - | - | - | - | - |
| | | Auto-Panoptic [73] | 44.8 | 35.0 | 51.4 | 78.9 | - | - | 54.5 | - | - |
| | | SPINet [103] | 42.2 | 31.4 | 49.3 | - | - | - | - | - | - |
| | | DR1Mask [104] | 46.1 | 35.5 | 53.1 | 81.5 | - | - | 55.3 | - | - |
| | | Gao et al. [107] | 45.7 | 37.5 | 51.2 | - | - | - | - | - | - |

[105], followed by 40.5% that has been obtained by EfficientPS [82] on val set. The PQ value of this dataset was almost similar to COCO. However, the SQ and RQ values of this dataset were lower than COCO. This is mainly due to the fact that COCO has more different, non-overlapping things compared to Mapillary Vistas. Table 6 summarizes the results obtained in other frameworks under Mapillary Vistas and Pascal VOC 2012 datasets.

On Mapillary Vistas val set and test-dev set, the obtained results are reported in Table 6. Generally from these results, the performance of the methods validated on this dataset describes a more challenging dataset compared to COCO and Cityscapes. This is because of the variation of the scenes, the number of objects (65 classes), as well as images in this repository. Explicitly, they have been captured during different seasons and weather conditions and during the day/night vision, which makes finding the pattern difficult for all the techniques. Even that, the PanopticDeepLab+ and EfficientPS methods have succeeded in the segmentation task with a high-performance rate versus the other methods. In terms of $PQ$, $PQ^{st}$ and $PQ^{th}$, PanopticDeepLab+ is more efficient with respect to the other methods. However, for the other metrics, EfficientPS has reached 74.9% in terms of $SQ$, while the fourth-best value of $SQ$ is 55.9%, which is attained

**Table 6**
Performance comparison of existing schemes on the val and test-dev sets under Mapillary Vitas, Pascal VOC 2012 and ADE20K datsets, where red, blue, cyan colors indicates the three best results for each set, respectively.

| Dataset | test/val | Method | PQ | | | SQ | | | RQ | | |
|---|---|---|---|---|---|---|---|---|---|---|---|
| | | | PQ | $PQ^{st}$ | $PQ^{th}$ | SQ | $SQ^{st}$ | $SQ^{th}$ | RQ | $RQ^{st}$ | $RQ^{th}$ |
| Mapillary Vistas | val | JSISNet [43] | 17.6 | 27.5 | 10 | 55.9 | 66.9 | 47.6 | 23.5 | 35.8 | 14.1 |
| | | EfficientPS Single-scale[82] | 38.3 | 44.2 | 33.9 | 74.2 | 75.4 | 73.3 | 48.0 | 54.7 | 43.0 |
| | | EfficientPS Multi-scale [82] | 40.5 | 47.7 | 35.0 | 74.9 | 76.2 | 73.8 | 49.5 | 56.4 | 44.4 |
| | | Panoptic-deeplab [39] | 40.3 | 49.3 | 33.5 | - | - | - | - | - | - |
| | | Geus et al. [40] | 23.9 | 35 | 15.5 | 66.5 | - | - | 31.2 | - | - |
| | | Mao et al. [77] | 38.3 | 44.4 | 33.6 | - | - | - | - | - | - |
| | | Porzi et al. [41] | 35.8 | 39.8 | 33.0 | - | - | - | - | - | - |
| | | PanopticDeepLab+ [105] | 44.3 | 51.9 | 38.5 | - | - | - | - | - | - |
| | test | Panoptic-deeplab [39] | 42.7 | 51.6 | 35.9 | 78.1 | 81.9 | 75.3 | 52.5 | 61.2 | 46.0 |
| | | Kirillov et al. [23] | 38.3 | 41.8 | 35.7 | 73.6 | - | - | 47.7 | - | - |
| Pascal VOC 2012 | val | FPSNet [67] | 57.8 | - | - | - | - | - | - | - | - |
| | | Zhang et al. [70] | 65.8 | - | - | - | - | - | - | - | - |
| | | Li et al. [92] | 63.1 | - | - | - | - | - | - | - | - |
| | | BCRF [88] | 71.76 | - | - | 89.63 | - | - | 79.33 | - | - |
| ADE20K | val | BGRNet [100] | 31.8 | 27.3 | 34.1 | - | - | - | - | - | - |
| | | PanopticDeepLab+ [105] | 37.4 | - | - | - | - | - | - | - | - |
| | | Ada-Segment[95] | 32.9 | 27.9 | 35.6 | - | - | - | - | - | - |
| | | Auto-Panoptic [73] | 32.4 | 30.2 | 33.5 | 74.4 | - | - | 40.3 | - | - |

using JSISNet [43]. This represents an improvement of 20%, which is a considerable difference.

Also, for all the other metrics, EfficientPS has achieved the highest performance rate due to the use of the multi-scale method of segmentation. On test-dev set, Panoptic-deeplab has reached the best value with 42.7% in terms of PQ, which demonstrate the efficiency of these type of methods using the Xception-71 backbone [39].

#### 6.2.4. Evaluation on Pascal VOC 2012

Pascal VOC 2012 is not the favorite dataset for testing panoptic segmentation strategies. By contrast to other datasets, e.g. Cityscapes and COCO that consist of stuff and things classes, PASCAL VOC 2012 encompasses thing classes as shown in Fig. 6, and hence the balance between stuff and thing classes is different than Cityscapes. However, it may be possible that changing hyperparameters can substantially improve performance. This demonstrates the performance of the frameworks presented in Table 6. For example, it has been shown that the performance of BCRF [71] has achieved 71.7%, which is the highest PQ result compared to other schemes evaluated on the same dataset as well as in comparison with other PQ results on other datasets, e.g. COCO and Cityscapes. Furthermore, the performance of BCRF has been more accurate than FPSNet by up to 14%. By and large, due to its specificity in comparison with the existing datasets, most of the panoptic segmentation frameworks have not trained their models on Pascal VOC 2020.

#### 6.2.5. Evaluation on ADE20K

The obtained results for each framework on ADE20K are reported in Table 6. On val set, the approach described using PanopticDeepLab+ has reached a performance rate of 37.4%, which outperforms Ada-Segement, Auto-Panoptic, and BSRNet in terms of PQ by more than 4%, 5% and 6%, respectively. Moreover, it has clearly been seen that this framework (PanopticDeepLab+) reports only PQ metrics, while the other evaluations on thing and stuff classes have not been considered (as illustrated in Table 6).

### 6.3. Evaluation using AP and mIoU metrics

As discussed previously, in addition to evaluating the quality of panoptic segmentation using PQ, RQ, and SQ, other solutions have only used AP and IoU metrics. To that end, we discuss in this section only the frameworks evaluated using AP and IoU metrics. Table 7 presents the obtained results of several existing panoptic segmentation works in terms of AP and IoU metrics with reference to different datasets, including Cityscapes, COCO, ADE20K, Mapillary Vitas, KITTI, and Semantic KITTI. From Table 7, it can be observed that PanopticDeepLab+, Panoptic-deeplab, and UPSNet are the best methods in terms of AP and IoU metrics under Cityscapes, COCO , Mapillary Vitas and ADE20K datasets. In this regard, for the case of Cityscapes and under the val set, the PanopticDeepLab+ method has reached the best AP value with a percentage of 46.8%, which is more accurate than the second best value by up to 34% that has been achieved using PanopticDeepLab+. On the other hand, under the COCO dataset, UPSNet has attained the highest mIoU value compared to SDC-Depth with an accuracy of 55.8%. For the remaining datasets, including Mapillary Vitas and ADE20K, PanopticDeepLab+ is the best technique in terms of the AP metric on the val set. On the cityscapes test set, PanopticDeepLab+ has also achieved the best results in comparison with Panoptic-deeplab as well as under Mapillary Vistas. Moving on, with regard to mIoU, PanopticDeepLab+ has reached higher performance results under Cityscapes (val set), Mapillary Vistas (val set), and ADE20K (for both val and test sets). While under the COCO dataset, it has been clearly seen that the SDC-Depth approach has achieved the best performance. In addition, under both KITTI and SemanticKITTI datasets, EfficientPS has attained the best result under both val and test set.

**Table 7**
Performance comparison using AP and IoU metrics

| Dataset | test/val | Method | AP | mIoU |
|---|---|---|---|---|
| Cityscpaes | val | EfficientPS Single-scale [82] | 38.3 | 79.3 |
| | | EfficientPS Multi-scale [82] | 43.8 | 82.1 |
| | | PanoNet [159] | 23.1 | – |
| | | CASNet[160] | 35.8 | - |
| | | Panoptic-deeplab [39] | 42.5 | 83.1 |
| | | PCV [80] | - | 74.1 |
| | | Hou et al. [98] | 29.8 | 77.0 |
| | | WeaklySupervised [92] | 24.3 | 71.6 |
| | | Porzi et al. [41] | 33.3 | 74.9 |
| | | UPSNet [46] | 39.0 | 79.2 |
| | | SDC-Depth [162] | - | 64.8 |
| | | PanopticDeepLab+ [105] | 46.8 | 85.3 |
| | | SPINet [103] | 35.3 | 80.0 |
| | | Chennupati et al. [106] | 24.9 | 68.7 |
| | | Petrovai et al [91] | - | 76.9 |
| | test | Panoptic-deeplab [39] | 39.0 | 84.2 |
| | | PanopticDeepLab+ [105] | 42.2 | 84.1 |
| COCO | val | SDC-Depth [162] | 31.0 | 38.6 |
| | | UPSNet [46] | 34.3 | 55.8 |
| | | SPINet [103] | 33.2 | 43.2 |
| Mapillary Vitas | val | Panoptic-deeplab [39] | 17.2 | 56.8 |
| | | EfficientPS Single-scale [82] | 18.7 | 52.6 |
| | | EfficientPS Multi-scale [82] | 20.8 | 54.1 |
| | | Mao et al. [77] | 17.6 | 52.0 |
| | | Porzi et al. [41] | 16.2 | 45.6 |
| | | PanopticDeepLab+ [105] | 21.8 | 60.3 |
| | test | Panoptic-deeplab [39] | 16.9 | 57.6 |
| ADE20K | val | PanopticDeepLab+ [105] | - | 50.35 |
| | test | PanopticDeepLab+ [105] | - | 40.47 |
| KITTI | val | EfficientPS Single-scale[82] | 27.1 | 55.3 |
| | | EfficientPS Multi-scale[82] | 27.9 | 56.4 |
| SemanticKITTI | val | EfficientLPS [131] | - | 64.9 |
| | | Panoster [128] | - | 61.1 |
| | test | EfficientLPS [131] | - | 61.4 |
| | | Panoster [128] | - | 59.9 |

## 6.4. Evaluation on LiDAR data

LiDAR refers to sensors used to calculate the distance between the source and the object using laser light and by calculating the time taken to reflect the light back. LiDAR sensor data is the mostly sought-after data when it comes to self-driving autonomous cars. On the other hand, panoptic segmentation is seen as a new paradigm that can help the self-driving cars by providing a finer grain and full understanding of the image scenes. To that end, different datasets have been launched to test the performance of the panoptic segmentation for LiDAR applications. For instance, SemanticKITTI, which is an extension of the KITTI dataset, a large-scale dataset providing dense point-wise semantic labels for all sequences of the KITTI Odometry Benchmark, has been employed for training and evaluating laser-based panoptic segmentation. Thus, SemanticKITTI is the benchmark dataset that is currently available for the evaluation of panoptic segmentation on LiDAR data.

Another aspect noticed in the literature is that both two-stage and unified based approach are carried out for panoptic segmentation on LiDAR data. A deep explanation of two-stage approaches can be found in [123], while for unified architectures, different sources can be consulted, including PanopticTrackNet+ [163], EfficientLPS [131], DSNet [132] and Panoster [128]. Moreover, a clustering based method can be seen in [127]. It is worth noting that for panoptic segmentation, some existing works have also performed ablation study. Furthermore, the authors of [163] have also made a pseudo-label annotation on a dataset after training and testing their model on the semanticKITTI dataset.

**Table 8**
Performance of various panoptic techniques under KITTI and SemanticKITTI.

| Dataset | test/val | Method | PQ | SQ | RQ | PQ$^{th}$ | SQ$^{th}$ | RQ$^{th}$ | PQ$^{st}$ | SQ$^{st}$ | RQ$^{st}$ |
|---|---|---|---|---|---|---|---|---|---|---|---|
| **KITTI** | val | EfficientPS Single-scale[82] | 42.9 | 72.7 | 53.6 | 30.4 | 69.8 | 43.7 | 52.0 | 74.9 | 60.9 |
| | | EfficientPS Multi-scale[82] | 43.7 | 73.2 | 54.1 | 30.9 | 70.2 | 44.0 | 53.1 | 75.4 | 61.5 |
| **SemanticKITTI** [122] | val | RangeNet++[126]+PointPillars | 37.1 | 47.0 | 75.9 | 20.2 | 25.2 | 75.27 | 49.3 | 62.8 | 76.5 |
| | | KPConv [125]+ PointPillars | 44.5 | 54.4 | 80.0 | 32.7 | 38.7 | 81.5 | 53.1 | 65.9 | 79.0 |
| | | PanopticTrackNet+ [130] | 40.0 | 73.0 | 48.3 | 29.9 | 33.6 | 76.8 | 47.4 | 70.3 | 59.1 |
| | | Milioto et al. [129]+ | 38.0 | 48.2 | 76.5 | 25.6 | 31.8 | 76.8 | 47.1 | 60.1 | 76.2 |
| | | EfficientLPS [131] | 65.1 | 75.0 | 69.8 | 58.0 | 78.0 | 68.2 | 60.9 | 72.8 | 71.0 |
| | | DSNet [132] | 63.4 | 68.0 | 77.6 | 61.8 | 68.8 | 78.2 | 54.8 | 67.3 | 77.1 |
| | | Panoster [128] | 55.6 | 79.9 | 66.8 | 56.6 | - | 65.8 | - | - | - |
| | test | EfficientLPS [131] | 63.2 | 83.0 | 68.7 | 53.1 | 87.8 | 60.5 | 60.5 | 79.5 | 74.6 |
| | | DSNet [132] | 62.5 | 66.7 | 82.3 | 55.1 | 62.8 | 87.2 | 56.5 | 69.5 | 78.7 |
| | | Panoster [128] | 59.9 | 80.7 | 64.1 | 49.4 | 83.3 | 58.5 | 55.1 | 78.8 | 68.2 |

Moreover, the val set and the testing set of SemanticKITTI dataset have been used to evaluate and study the performance of the existing methods, as portrayed in Table 8. *PQ*, *SQ* and *RQ* for both "stuff" and "things" are reported for better clarification. In this context, EfficientLPS [131] achieves a *PQ* score of 65.1%, which has an improvement of 4.7% over the previous state-of-the-art Panoster. EfficientLPS has achieved the highest *PQ* value in both the test and the val set of semanticKITTI. Not only in terms of *PQ*, but also the values of *SQ* and *RQ* for both "stuff" and "things" are the highest for EfficientLPS. Therefore, EfficientLPS has an increase of 4.5% in terms of the *SQ$^{th}$* score compared to Panoster. This is most probably because of the use of the peripheral loss function, which improves the SQ of has been used. Similarly, distance-dependent semantic head has been used to improve the *RQ$^{st}$* by 2% and an overall *RQ* score of 69.8%.

Moving on, DSNet [132] has achieved better *PQ* and *RQ* performance than Panoster, however, panoster has better *SQ* results than DSNet. Apart from the *PQ*, *SQ* and *RQ* metrics, mIoU has also been used in the evaluation, as presented in Table 8. The highest value of 64.9% has been achieved by EfficientLPS on the val set of semanticKITTI while Panoster got 61.1%. Similarly, on the test set also EfficientLPS has better performance than Panoster. It can be concluded that out of the available panoptic segmentation methods for LiDAR, the EfficientLPS scheme has the best performance and can be considered as the current state-of-the-art method.

### 6.5. Evaluation on medical dataset

To compare the effectiveness of proposed methods for medical microscopy images, fluorescence and histopathology microscopy datasets have been used, such as BBBC039V1, Kumar, and TNBC. BBBC039V1 is a fluorescence microscopy image dataset of U2OS cells. Kumar and TNBC are histopathology microscopy datasets. Some frameworks have used the BBBC039V1 dataset as the training dataset and either Kumar or TNBC as as the target dataset.

Table 9 represents the performance of each method using the two scenarios. Panoptic SQ has been used here to demonstrate the effectiveness of different methods as well as two other metrics, including AJI and F1 metrics. By considering the false-positive predictions, AJI extends the Jaccard Index for each object, which multiplies the F1 score for object detection and the IoU score for instance segmentation. The methods that have trained their model using the same scenarios include CYCADA [164], Chen el al. [165], SIFA [166], DDMRL [167], Hou et al. [168], PDAM [36] and Cyc-PDAM [111].

From the obtained results, we can find that the PDAM method has reached the highest values in terms of AJI and F1 metrics better than CYC-PDAM, therefore up to 0.4% and 1% improvements in terms of AJI and F1 have been obtained, respectively, under BBC039(Kumar) dataset. In addition, under BBC039(TNBC), PDAM has been more accurate than CYC-PDAM by 1% for the AJI metric and 2% in terms of the F1 metric. Moreover, it has outperformed the results of other methods by more than 10% for AJI, such as SIFA and DDMR, while up to 7% improvement has been attained comapred to Hou el a., DDMRL, and SFIA using F1 score. Regarding the panoptic SQ and PQ, Cyc-PDAM has been the best accurate method compared to PDAM or other methods. Accordingly, Cyc-PDAM has been better than PDAM by up to 22% on BBC039(Kumar) scenario, while up to 20% improvement has been reached under the BBC039(TNBC) scenario.

## 7. Challenges and Future Trends

### 7.1. Current Challenges

As explained previously, panoptic segmentation is a combination of semantic and instance segmentation while semantic segmentation is the contextual pixel-level labeling of a scene and instance segmentation is the labeling of the objects contained in this scene. For semantic-based, the categorization of a pixel made by deciding which class this pixel belongs to, where the instance classification exploits the results of object detection followed by a fine-level segmentation to labeling the object pixels in one homogeneous label.

**Table 9**
Comparison of the performance on both two histopathology and fluorescence microscopy datasets.

| Method | BBBC039 $\rightarrow$ *Kumar* | | | BBBC039 $\rightarrow$ *TNBC* | | |
|---|---|---|---|---|---|---|
| | AJI | F1 | PQ | AJI | F1 | PQ |
| CyCADA [164] | 0.4447±0.1069 | 0.7220±0.0802 | 0.6567±0.0837 | 0.4721±0.0906 | 0.7048±0.0946 | 0.6866±0.0637 |
| Chen et al. [165] | 0.3756±0.0977 | 0.6337±0.0897 | 0.5737±0.0983 | 0.4407±0.0623 | 0.6405±0.0660 | 0.6289±0.0609 |
| SIFA [166] | 0.3924±0.1062 | 0.6880±0.0882 | 0.6008±0.1006 | 0.4662±0.0902 | 0.6994±0.0942 | 0.6698±0.0771 |
| DDMRL [167] | 0.4860±0.0846 | 0.7109±0.0744 | 0.6833±0.0724 | 0.4642±0.0503 | 0.7000±0.0431 | 0.6872±0.0347 |
| Hou et al. [168] | 0.4980±0.1236 | 0.7500±0.0849 | 0.6890±0.0990 | 0.4775±0.1219 | 0.7029±0.1262 | 0.6779±0.0821 |
| CyC-PDAM [111] | 0.5610±0.0718 | 0.7882±0.0533 | 0.7483±0.0525 | 0.5672±0.0646 | 0.7593±0.0566 | 0.7478±0.0417 |
| PDAM [36] | 0.5653±0.0751 | 0.7904±0.0474 | 0.5249±0.0884 | 0.5726±0.0414 | 0.7742±0.0302 | 0.5409±0.0331 |

The semantic segmentation can include the segmentation of the stuff and things together while the things are labeled with the same color class correspond to the object type. While the instance segmentation gives the separation of these objects using different color classes. Similar to all computer vision tasks, many challenges can obstruct any method to reach the best outcomes. In this perspective, different limitations have been identified, such as the occlusion between the objects, scale variation of the objects, illumination changes and last but not least the similar intensities of the objects. To that end, in this paper we attempt to summarize some of the challenges that are currently faced as follows:

- **Objects scale variations:** It is one of the limitations for all computer vision tasks, including objects detection, semantic, instance and panoptic segmentation. Most of the proposed models attempt to address this problem as a first step. Generally, existing methods do not work well on small targets, while available annotated datasets for training are not sufficient in terms of scenes that include many scaled objects [169, 170]. Detection of small objects within images is quite difficult, and furthermore distinguishing them into things and stuff is even more harder when the object is small, especially when images are skewered and occluded.

- **Complex background:** For image segmentation, many (stuff, things) can be considered as other annotated (stuff, things) when scenes are complex. The captured images can include many (stuff, things) that are not annotated in the datasets, which makes the appearance of people and other objects similar [171].

- **Cluttered scenes:** The total or partial occlusion between dynamic objects in the scene is also one of the limitations of most of the panoptic segmentation methods. This is especially for the case of instance (things) segmentation which is an essential part in panoptic segmentation that can be affected by the occlusion . Thus, this results in much lower quality and quantity of the 'things' segmented.

- **Weather changes:** The surveillance using drones can be affected by various types of weather conditions and

environmental changes such as rain, could, and fog. Accordingly, this could reduce the accuracy of panoptic segmentation algorithms once they are applied in real-world scenarios [172].

- **Quality of datasets:** It is highly important for improving the performance of panoptic segmentation models. Although there are several available datasets, there is still a struggle in annotating them [173, 174, 175]. While panoptic segmentation and segmentation in general require data to be annotated or validated by human experts.

- An efficient merging heuristic is required to merge the instance and semantic segmentation results and produce the final panoptic segmentation visual results. The accuracy of the merging heuristic determine the performance of the model in general. However, a critical problem in this case is due to the increase in computation time because of the merging heuristic algorithm.

- **Computational time:** The training time using DL models for panoptic segmentation is generally very costly due to the complexity of these models, and also because of the nature of the model, i.e. single or separated. Generally speaking, separated model (instance semantic for panoptic) takes more training time than unified models, however, the panoptic SQ is better.

### 7.2. Future Trends

In the near future, more research works can be concentrated on developing end-to-end models to perform both instance and semantic segmentation simultaneously. This will reduce the demand for merging heuristics, as merging also weighs in as a factor for measuring the performance of the model. Replacing the merging heuristic methods can improve further the computational time for the model [67].

More concentration on detecting smaller objects, removal of unnecessary small objects and other miscellaneous objects can be given. Also the separation between things for a good instance segmentation can be helpful using accurate edge detection methods. This will also help to provide some real-time panoptic segmentation techniques. At the moment, there is only a very limited number of real-time applications of the panoptic segmentation deployed so far. Hence, it will be of utmost importance to focus on this perspective in the future. Moreover, improving the performance of panoptic segmentation models and widening their applications are among the relevant future orientations, especially in digital health, real-time self-driving, scene reconstruction and 3D/4d Point Cloud Semantic Segmentation, as it is explained in the following subsections.

### 7.2.1. Medical imagery

A great hope is put on panoptic segmentation to improve medical image segmentation in the near future. Indeed, panoptic segmentation of the amorphous regions of cancer cells from medical images can help the physicians to detect and diagnose disease as well as the localization of the tumors. This is because the morphological clues of different cancer cells are important for pathologists for determining the stages of cancers. In this regard, panoptic segmentation could help in obtaining the quantitative morphological information, as it is demonstrated in [112], where an end-to-end network for panoptic segmentation is proposed to analyze pathology images. Moreover, while most of existing cell segmentation methods are based on the semantic-level or instance-level cell segmentation, panoptic segmentation schemes unify the detection and localization of objects, and the assignment of pixel-level class information to regions with large overlaps, e.g. the background. This helps them in outperforming the state-of-the-art techniques.

### 7.2.2. Real-time self-driving

Self-driving becomes one of the bronzing advances due to its impact on daily lives as well as on urban planing and transportation techniques. This has encouraged the researchers to tackle different challenges for increasing the performance of self-driving cars during the last teen years. With the existing techniques especially AI, e.g. neural networks and DL have contributed to overcome many limitations of autonomous driving. Using these techniques combined with different sensors including cameras and LiDARs helps for scene understanding and object localization that represent crucial tasks for self-driving [176]. Also by knowing and localizing the objects around the car as well as the surface that the car driving on, the driving safety can be ensured [91].

In this context, panoptic segmentation can significantly contribute to the identification of these objects (thing), e.g. to read signs and detect people crossing the roads, especially on busy streets [177], in addition to the segmentation of the driving roads (stuff) [178]. Fig. 7 illustrates an example of the applicability of panoptic segmentation for autonomous vehicles. This is also possible by using adequate computing boards that enable the training of panoptic segmentation models based on DL for better understanding the scene as a whole versus piecewise.

However, when panoptic segmentation is used for self-driving it is required to be real-time. Also, as it is often combined with DL models, it becomes a crucial challenge to develop real-time self-driving solutions based on panoptic segmentation. Therefore, a great effort should be put in this line in the near future to to continue the studies recently launched. For instance, NVIDIA has developed, DRIVE AGX[13], which is a powerful computing board that helps in simultaneously running the panoptic segmentation DNN along with many other DNN networks and perception functions, localization, and planning and control software in real-time.

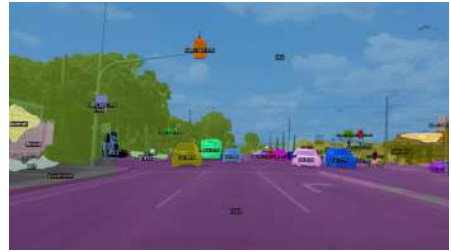

**Figure 7:** Example of using panoptic segmentation in autonomous driving.

### 7.2.3. Scene reconstruction

Real-time dynamic scene reconstruction is one of the hot topic in visual computing. its benefit can be found on real-world scenes understanding also on all current applications including virtual reality, robotics, etc. Using 3D-based sensors, such as LiDARs, or cameras data the scene reconstruction becomes easier with deep learning techniques. Existing multi-view dynamic scene reconstruction methods either work in controlled environment with known background or chroma-key studio or require a large number of cameras [179],[180]. Panoptic segmentation can made a crucial improvement on scene reconstruction methods, due to the simplification of the complex scene as well as the separation using color classes that make the understanding of the context of a scene then an accurate reconstruction of it, as it is shown in Fig. 8 [181]. Exploiting panoptic segmentation on 3D LiDAR data also makes this reconstruction easier for 3D shapes which is more similar to real-world scenes.

### 7.2.4. 3D/4d Point Cloud Semantic Segmentation

3D/4D Point Cloud Semantic Segmentation (PCSS) is a cutting-edge technology that attracts a growing attention because of its application in different research fields, such as computer vision, remote sensing, and robotics, thanks to the new possibilities offered by deep neural networks. 3D/4D PCSS refers to the 3D/4D form of semantic segmentation, where regular/irregular distributed points in 3D/4D space are employed instead of regular distributed pixels in a 2D image. However, in comparison with the visual grounding in 2D images, 3D/4D PCSS is more challenging due to the sparse and disordered property. To that end, using panoptic segmentation can efficiently improve the performance of 3D/4D PCSS. Accordingly, based on the predicted target category from natural language, the authors in [182] have proposed panoptic based model, namely InstanceRefer, to firstly filter instances from panoptic segmentation on point clouds for obtaining a small number of candidates. Following, they conducted the cooperative holistic scene-language understanding for every candidate before localizing the most pertinent candidate using an adaptive confidence fusion. This has helped InstanceRefer in effectively outperforming existing techniques. Moving on, since temporal semantic scene analysis is essential to develop powerful models for self-driving vehicles and even autonomous robots that operate in dynamic environments, Aygun et al. [133] introduce a 4D panoptic segmentation for assigning a semantic class and a temporally consistent instance ID to a sequence of 3D points. Typically, this method has relied on determining a semantic class for each point and modeling object instances as probability distributions in the 4D spatio-temporal domain. Therefore, multiple point clouds have been processed simultaneously and point-to-instance associations have been resolved. This allows to efficiently alleviate the need for explicit temporal data association.

---

[13] https://www.nvidia.com/en-us/self-driving-cars/drive-platform/hardware/

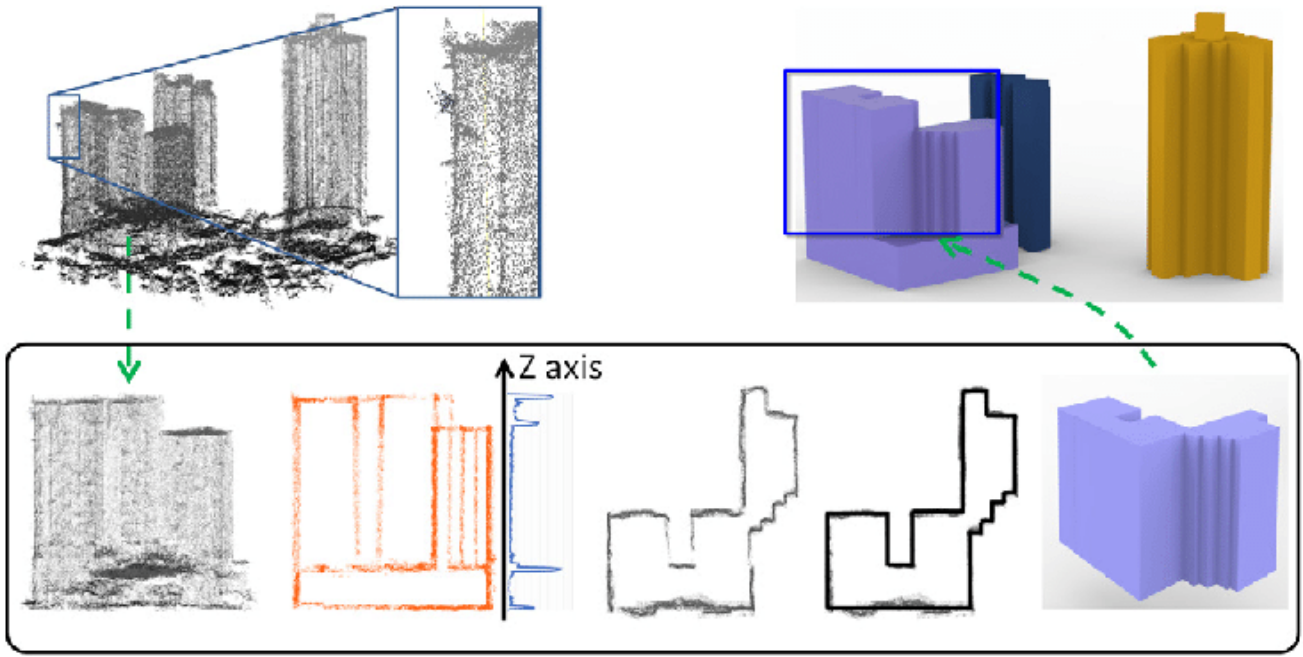

**Figure 8:** Example of 3D scenes reconstruction using LiDAR point cloud [181].

## 8. Conclusion

Panoptic segmentation is a breakthrough in computer vision that segments "things" and "stuff" by separating objects into different classes. Panoptic segmentation has opened several opportunities in various research and development fields. A separation of things and stuff with the distinction of objects is required, such as autonomous driving, medical image analysis, remote sensing images mapping, etc. To inform the state-of-the-art, we proposed the first extensive critical review of the panoptic segmentation technology to the best of the authors' knowledge, which has been designed following a well-defined methodology. Accordingly, the background of the panoptic segmentation technology was first presented. Next, a taxonomy of existing panoptic segmentation schemes was performed based on the nature of the adopted approach, type of image data analyzed, and application scenario. Also, datasets and evaluation metrics used to validate panoptic segmentation frameworks were discussed, and the most pertinent works were tabulated for a clear comparison of the performance of each model.

In this context, it was clear that some methods performed instance segmentation and semantic segmentation separately and fused the results to achieve panoptic segmentation, while most of the existing techniques completed the process as a unified model. Although, the significant importance given to the panoptic segmentation by the research community has led to the publication of various research articles. The PQ of 69% on the Cityscapes dataset and 50% on the COCO dataset were the best results from all the models. This demonstrates that significant work is still required to improve their performance and facilitate their implementation.

In terms of the applications of panoptic segmentation, there was a strong inclination towards autonomous driving, pedestrian detection, and medical image analysis (especially using histopathological images). However, new application opportunities are emerging, such as for the military sector, where panoptic segmentation can be utilized to visualize hidden enemies on battlefields. On another side, although there are very few real-time applications of panoptic segmentation yet, an increasing interest is devoted towards this direction. One of the most striking features of panoptic segmentation is its ability to annotate datasets, which significantly reduces the computation time required for the annotation process.

## Acknowledgments

This research work was made possible by research grant support (QUEX-CENG-SCDL-19/20-1 ) from Supreme Committee for Delivery and Legacy (SC) in Qatar.